\documentclass{article} 
\usepackage{iclr2027_conference,times}

\usepackage{amsmath,amsfonts,bm}

\def\eqref#1{equation~\ref{#1}}

\def\1{\bm{1}}

\DeclareMathAlphabet{\mathsfit}{\encodingdefault}{\sfdefault}{m}{sl}
\SetMathAlphabet{\mathsfit}{bold}{\encodingdefault}{\sfdefault}{bx}{n}

\usepackage{hyperref}
\usepackage{url}
\usepackage{booktabs}
\usepackage{graphicx}
\usepackage[table]{xcolor}
\usepackage{array}
\usepackage{multirow} 
\usepackage{adjustbox}
\usepackage{makecell}
\usepackage{subcaption}
\usepackage[table]{xcolor}
\usepackage{caption}
\usepackage[normalem]{ulem} 

\usepackage{tabularx}

\title{HiUni: Harmonizing Understanding and Generation in Shared Visual Space}

\iclrfinalcopy
\author{
ERNIE Team, Baidu Inc. \\
{
\raisebox{-0.16em}{\includegraphics[height=0.9em]{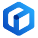}}\,
{Project Page:
\url{https://ernie-research.github.io/HiUni}
}
}
}

\newcommand{\hiuni}{\textsc{HiUni}}
\begin{document}

\maketitle
\lhead{Preprint}        

\begin{abstract}
Unified multimodal models aim to support visual understanding and generation within a single framework.
However, prevailing approaches often separate their visual representations or backbone computation to accommodate the different requirements of these two tasks, which limits the degree of unification.
In this paper, we present \hiuni, which revisits this design choice and explores whether a pretrained vision language model can be directly extended into a unified model with extensive parameter sharing.
Specifically, \hiuni{} reuses the same language-aligned ViT and dense Transformer backbone for both capabilities, while confining task-specific computation to the output side.
Yet such a design introduces two key challenges.
Generation in pixel space is difficult to optimize, and semantic abstraction in pretrained ViT discards fine spatial details required for synthesis.
To address these challenges, \hiuni{} introduces a \emph{Pixel Diffusion Head} tailored to direct pixel-space optimization and a lightweight \emph{HiGate} pathway that provides fine-grained spatial details near the output stage.
Extensive experiments show that \hiuni{} achieves a similar average understanding score to the pretrained VLM while attaining competitive performance on image generation and editing.
Controlled comparisons and analyses further demonstrate that a single dense backbone can effectively support both capabilities without task-specific backbone computation, highlighting a simpler path toward unified multimodal modeling.
\end{abstract}



\section{Introduction}
\label{sec:introduction}

Image understanding translates visual content into language, whereas visual generation synthesizes images from textual descriptions.
An ideal unified multimodal model (UMM) is expected to support both capabilities within a single framework and facilitate cross-task interaction between them~\citep{kang2026transferability,wu2026uncovering}.
However, most existing UMMs still maintain two separate visual representation spaces with dedicated encoders.
The understanding branch extracts high-level semantics via a language-aligned Vision Transformer~(ViT), while the generation branch learns fine-grained spatial details with a Variational Autoencoder (VAE).
As highlighted by Janus~\citep{wu2025janus}, forcing two divergent encoding objectives into one representation triggers severe task interference.
Bagel~\citep{deng2025emerging} further introduces a Mixture-of-Transformers~(MoT) architecture that separates task-specific computation within the backbone, establishing a modular paradigm adopted by recent studies~\citep{agarwal2026cosmos,diao2026sensenova,yan2026unispace}.


Although this separation preserves individual performance, it limits direct interaction through shared visual representations.
Recent studies further suggest that closer sharing of visual representations can promote stronger interaction between the two capabilities with an appropriate backbone design~\citep{zheng2025aia,su2026generation}.
These observations motivate us to revisit whether such architectural separation is indeed necessary:
\emph{Can a shared visual space internally harmonize understanding and generation without relying on task-specific backbone computation?}

Recent advances offer promising opportunities to address this question using a single visual encoder for both understanding and generation~\citep{liu2026tuna,zhang2026hydra}.
Conventional generation models heavily rely on VAE latents, yet these features are optimized primarily for reconstruction and are less aligned with the high-level semantics required for understanding~\citep{wu2025vila,ma2026unitok}.
Conversely, modern pixel-space diffusion models demonstrate that high-quality generation can also be achieved without VAE latents~\citep{li2026back,yu2026pixeldit}, opening a new path towards shared visual representations.
Going one step further, some encoder-free architectures process raw pixels for both understanding and generation~\citep{liu2026tuna2,diao2026sensenova}, but the absence of pretrained visual semantic priors makes optimization substantially less stable and less data efficient.
An alternative choice is to reuse language-aligned ViT as the shared visual encoder, given its strong semantic representations and direct compatibility with standard Vision Language Model~(VLM) infrastructures.
Nevertheless, adapting such an architecture to image generation introduces a spatial bottleneck.
VLMs typically employ a patch merger after ViT to compress visual tokens, doubling effective patch size from $14/16$ to $28/32$~\citep{bai2025qwen3vl,wang2025internvl35}.
Such aggressive spatial pooling, together with the semantic abstraction of deep ViT layers, improves understanding efficiency, but weakens fine-grained information needed for high-fidelity synthesis.

Following this direction, we introduce \hiuni, a UMM initialized from a pretrained VLM that extends image understanding to generation through a shared semantic visual pathway.
Understanding and generation use the same visual encoder, token representation, and dense backbone.
Task-specific computation is confined to the output modules. 
For image generation, we observe optimization instability when applying a standard DiT to pixel-space diffusion in our early experiments.
We therefore design a dedicated \emph{Pixel Diffusion Head} with a simplified modulation for pixel-space generation.
To compensate for spatial information weakened by patch merging and deep semantic abstraction, we incorporate a lightweight gating module (\textit{HiGate}), inspired by Cheers~\citep{zhang2026cheers}, which directly routes uncompressed pixel-level features to the pixel diffusion head.
This residual pathway supplies local spatial cues for image generation while leaving semantic modeling in the shared visual encoder and backbone.

\begin{figure*}[t]
    \centering
    \includegraphics[
        width=1.0\textwidth
    ]{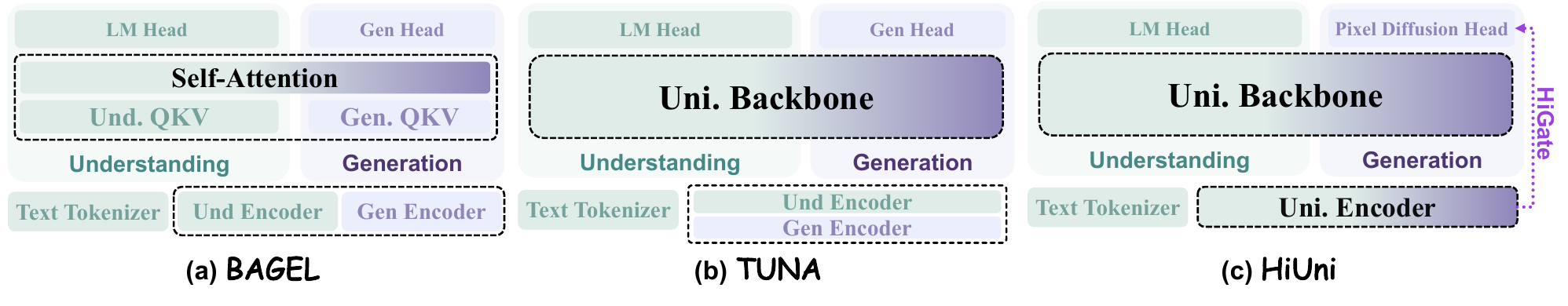}
    \caption{Comparison of representative UMM designs. 
    (a) Bagel utilizes separate visual encoders and task-specific MoT modules for understanding and generation. 
    (b) TUNA constructs a shared visual representation by cascading VAE and ViT.
    (c) \hiuni{} reuses the pretrained VLM visual pathway, while \emph{HiGate}  supplies fine-grained pixel details only to the generation head.}
    \label{fig:semantic_organization}
\end{figure*}

We comprehensively validate \hiuni{} on image understanding, image generation, and image editing.
\hiuni{} retains strong visual understanding capabilities inherited from the pretrained VLM while incorporating image generation into the shared backbone.
It also achieves the highest overall Chinese score on Qwen-Image-Bench among the compared UMMs, together with competitive image editing performance.
More importantly, further architectural comparison  reveals the trade-offs between shared and task-specific backbone computation. While task-routed MoT improves generation over a same-scale dense backbone under matched training settings, a larger shared dense backbone with a comparable total parameter count achieves substantially stronger understanding while maintaining competitive generation. These results suggest that scaling a shared dense backbone within a shared visual space offers a simple and promising path toward unified multimodal modeling.

Our contributions are summarized as follows:
\begin{itemize}
    \item We propose \hiuni, which provides a new perspective on building UMMs from pretrained VLMs.  Understanding, generation, and editing share the same language-aligned encoder and dense backbone, while task-specific computation is limited to the output module.
    \item We develop a \emph{Pixel Diffusion Head} tailored to direct pixel-space generation, together with a lightweight \emph{HiGate} pathway that complements the shared semantic representation with fine-grained spatial details.
    \item Extensive experiments show that \hiuni{} maintains strong understanding while achieving competitive generation and editing performance. 
    Controlled comparisons further highlight that a shared dense backbone has the potential to effectively support both understanding and generation without task-specific backbone computation.
\end{itemize}

\section{Related Work}
\label{sec:related_work}

\noindent \textbf{Unified Multimodal Models.}
UMMs are intended to support visual understanding and generation within a single model.
Existing UMMs mainly follow two paradigms according to how images are represented and generated.
Autoregressive models tokenize images into discrete visual tokens and model text and vision under a unified autoregressive objective~\citep{team2024chameleon,cui2025emu35,wu2025vila,team2026longcatnext,wang2026ernie5}.
This formulation provides a simple and scalable interface across modalities, but discrete tokenization may lose subtle visual details required for high fidelity image synthesis.
Another line of work follows the Transfusion paradigm~\citep{zhou2025transfusion}, which retains the autoregressive prediction of language while applying diffusion or flow matching to continuous visual features for image generation~\citep{deng2025emerging,liu2026tuna,yan2026unispace}.
More recently, pixel-space diffusion further demonstrates that strong image synthesis can be achieved without VAE latents~\citep{li2026back,yu2026pixeldit}, 
and local details can be recovered from noisy pixels to complement global semantic modeling~\citep{chen2026dip,ma2026deco}, opening a more direct path toward shared visual representations.
Together, these advances shift an important architectural question toward how visual representations and backbone computation should be shared between understanding and generation.


\noindent \textbf{Visual Unification.}
The architectural design of UMMs depends on how visual representations and backbone computation are organized.
Many existing models maintain separate visual spaces due to the different requirements of understanding and generation, where a language-aligned ViT provides semantic features for understanding, and a reconstruction-oriented VAE preserves fine-grained details for generation~\citep{wu2025janus,deng2025emerging}.
To reduce this separation, recent studies explore more unified visual pathways.
TUNA cascades a VAE encoder with a representation encoder to construct a shared visual representation, while TUNA-2 further removes both encoders for pixel space modeling~\citep{liu2026tuna,liu2026tuna2}.
Cheers decouples semantic representations and local details through a unified visual tokenizer and gated detail residuals in generation head~\citep{zhang2026cheers}.
HYDRA adopts a progressive ViT, and UniSpace introduces reconstruction aware patch representations into a pretrained ViT~\citep{zhang2026hydra,yan2026unispace}.
These designs reduce visual separation, but differ in their use of pretrained visual encoders and task-specific backbone computation.
For example, UniSpace still adopts task-specific Transformer experts, while encoder-free models such as SenseNova remove both ViT and VAE but retain an MoT backbone~\citep{yan2026unispace,diao2026sensenova}.
In another related direction, RepFusion processes noisy visual representations through a pretrained VLM, but relies on a separate diffusion Transformer for representation-space generation~\citep{pan2026repfusion}.
\hiuni{} reuses the ViT and patch merger of a pretrained VLM and shares their output representation and a single dense backbone across understanding and generation, without separate generative visual encoders or task-specific backbone computation.

\section{Method}
\label{sec:method}


Image understanding and generation differ substantially in their input and output forms, but both rely on semantic alignment between visual content and language.
\hiuni{} therefore separates shared semantic modeling from task-specific prediction.
Specifically, we seek to let the visual encoder and multimodal backbone provide a common semantic pathway for both capabilities, while introducing generation-specific processing only when detailed visual signals need to be reconstructed.
This design allows understanding and generation to share the visual and backbone computation without requiring separate semantic representations.

\begin{figure*}[t]
    \centering
    \includegraphics[width=1.0\textwidth]{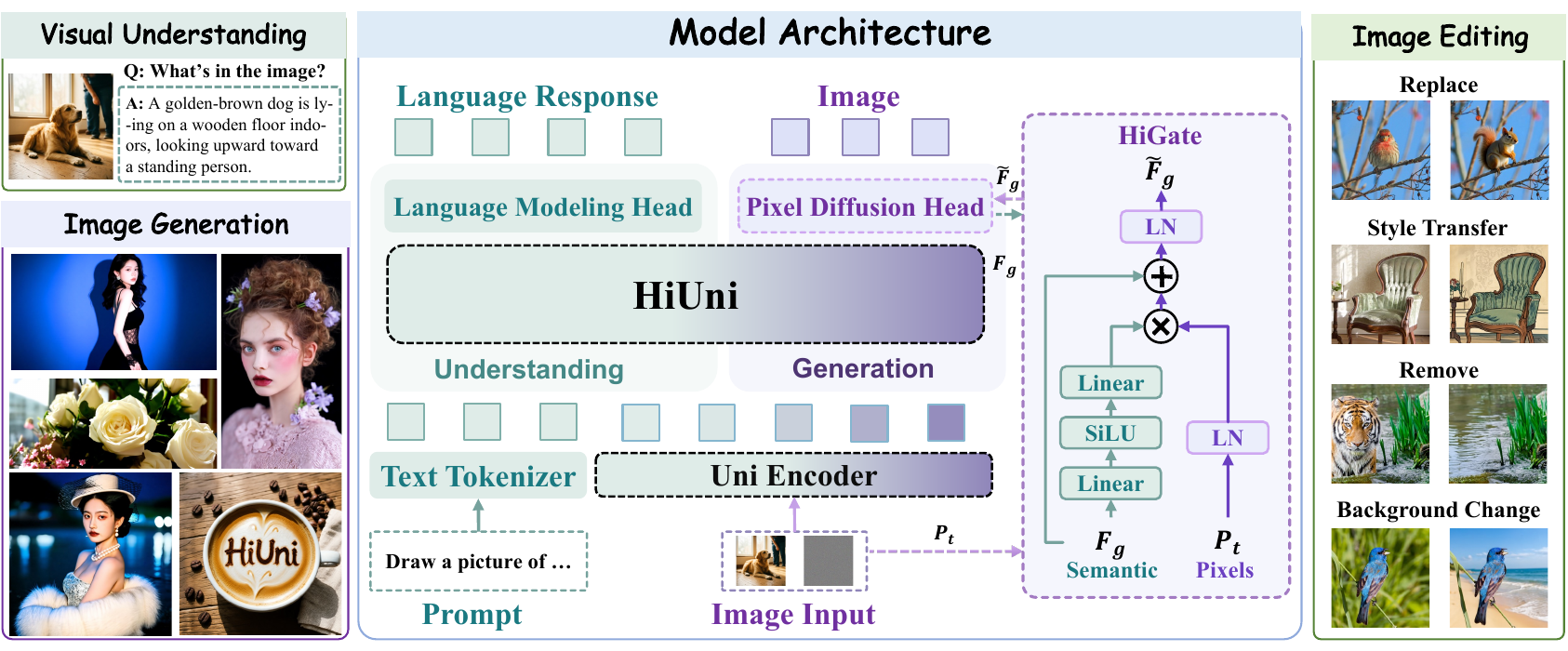}
    \caption{
        Overview of \hiuni. 
        Starting from a pretrained VLM, \hiuni{} shares the same visual encoder and dense backbone across understanding, generation, and editing. Language understanding is handled by a language modeling head, while image synthesis is performed by a \emph{Pixel Diffusion Head}. To bridge semantic representations and pixel-level details, \emph{HiGate} provides a lightweight residual pathway that supplies local spatial cues weakened by visual compression. 
    }
    \label{fig:framework}
\end{figure*}

Following this principle, \hiuni{} is built upon a pretrained VLM and extends it with image generation capability while reusing its pretrained visual encoder and backbone architecture.
Both understanding and generation use the same language-aligned ViT, the same visual tokens after patch merging, and the same dense backbone.
Textual outputs are predicted by the native language modeling head, while image synthesis is handled by a new \emph{Pixel Diffusion Head}. 
No separate semantic visual encoder or task-specific backbone is involved.
To compensate for the spatial information weakened by patch merging and deep semantic abstraction, \hiuni{} further adopts a lightweight gating module, termed \emph{HiGate}, which routes raw features from the noisy image to the diffusion head.
The residual branch supplies local spatial cues to complement the shared semantic representation without modifying the main semantic computation.
Figure~\ref{fig:framework} illustrates the overall architecture.

\subsection{Unified Multimodal Architecture}
\label{sec:unified_architecture}

\paragraph{Shared Visual Encoding.}
\hiuni{} adopts a language aligned ViT from the pretrained VLM as the only visual encoder for both understanding and generation.
Let $\mathcal{E}$ denote the ViT and $\mathcal{M}$ denote the patch merger, which compresses the ViT outputs into visual tokens.
For an input image $\mathbf{x}$, the visual tokens are obtained as $\mathbf{Z}=\mathcal{M}(\mathcal{E}(\mathbf{x}))$.
The same pathway is used for clean images in understanding and visual conditioning, as well as noisy target images in generation.
Consequently, both tasks are grounded in the same visual token space without requiring separate or task-specific encoders.

\paragraph{Shared Dense Backbone.}
The shared visual tokens are further processed by the same dense backbone for both understanding and generation.
Text embeddings and visual tokens are organized into a multimodal sequence $\mathbf{S}$ and processed by a single dense Transformer $\mathcal{B}$, yielding $\mathbf{H}=\mathcal{B}(\mathbf{S})$.
Attention mask allows bidirectional interaction within each image, while text tokens retain causal attention for autoregressive prediction.
After the shared backbone, task-specific prediction is handled on the output side.
For image understanding, the language modeling head projects hidden states corresponding to textual outputs onto the vocabulary space.
For image generation, hidden states at the target visual positions are passed to the pixel diffusion head for image prediction.

\subsection{Pixel-Space Image Generation}
\label{sec:visual_generation}

\paragraph{Pixel Diffusion Head.}
Diffusion directly in pixel space poses additional optimization challenges compared with conventional latent diffusion, as the model must predict high dimensional pixel signals while maintaining semantic consistency and fine spatial structure~\citep{jiang2026empirical,li2026back}.
Consistent with these observations, we observe frequent loss spikes when using a standard DiT~\citep{peebles2023scalable} with repeated AdaLN modulation in our early experiments.
In subsequent development runs, a simplified design without block-wise AdaLN exhibits smoother training trajectories.
Motivated by this observation, we remove AdaLN from the DiT blocks, reducing the diffusion head to a standard Transformer architecture.

Timestep conditioning enters through an explicit time token and a final modulation layer.
Inspired by recent works~\citep{ma2025janusflow,team2025hunyuanimage},
for timestep $t$, a sinusoidal encoding $\phi(t)$ is mapped by a two layer MLP to a time embedding $\mathbf{e}_t\in\mathbb{R}^{D_b}$, where $D_b$ denotes the hidden dimension of the backbone.
The resulting vector replaces the embedding of the \texttt{<time>} token and is processed together with the multimodal sequence by the shared backbone.
The backbone hidden states for the target visual positions, together with the time token representation, are projected to the diffusion hidden dimension $D$ and processed by $N$ Transformer blocks.

At the output of the diffusion head, we retain a single timestep modulation before pixel prediction.
A separate two layer MLP maps the same sinusoidal encoding $\phi(t)$ to a diffusion time embedding $\mathbf{c}_t\in\mathbb{R}^{D}$.
This embedding is further projected to channel wise scale and shift parameters as $[\boldsymbol{\gamma}_t,\boldsymbol{\beta}_t]=\mathbf{W}_{\mathrm{mod}}\operatorname{SiLU}(\mathbf{c}_t)+\mathbf{b}_{\mathrm{mod}}$.
Let $\mathbf{F}_{\mathrm{out}}\in\mathbb{R}^{N_g\times D}$ denote the target visual features at the output of the diffusion head.
The final modulation is given by
$\bar{\mathbf{F}}_{\mathrm{out}}=(1+\boldsymbol{\gamma}_t)\odot\operatorname{LN}(\mathbf{F}_{\mathrm{out}})+\boldsymbol{\beta}_t$.
After modulation, a linear projection predicts the corresponding pixel patches.
With a ViT patch size of $16$ and a spatial merge factor of $2$, each generated token predicts a $32\times32$ RGB region, corresponding to a flattened patch dimension of $N_p =3\times32\times32$.
Finally, the predicted patches are rearranged into image space to obtain $\hat{\mathbf{x}}_0$.
In this way, the head removes repeated block-wise AdaLN while preserving timestep conditioning through the shared backbone and final modulation layer.

\paragraph{HiGate.}
Although the unified visual tokens carry strong semantics, the patch merger reduces their spatial resolution, while deeper ViT layers increasingly favor abstraction over local texture.
Such compression is well suited to understanding but may discard cues important for pixel-level reconstruction.
To complement the semantic pathway, \hiuni{} introduces a lightweight \textit{HiGate} module, inspired by Cheers~\citep{zhang2026cheers}, which directly injects low-level pixel information into the diffusion head without introducing an additional visual encoder.

Specifically, the noisy image $\mathbf{x}_t\in\mathbb{R}^{3\times H\times W}$ branches into two pathways before ViT patch embedding.
The main pathway encodes $\mathbf{x}_t$ through the ViT, patch merger, and the shared backbone to obtain semantic features, while a lightweight residual branch directly preserves its local pixel content.
In the residual branch, each $2\times2$ group of $16\times16$ RGB patch is concatenated into $\mathbf{R}_t\in\mathbb{R}^{N_g\times N_p}$, following the same spatial layout as the target visual tokens.
A linear projection maps these raw pixel features to the diffusion hidden dimension $D$, producing $\mathbf{P}_t\in\mathbb{R}^{N_g\times D}$.
The aligned pixel features $\mathbf{P}_t$ are then fused with the target features $\mathbf{F}_g\in\mathbb{R}^{N_g\times D}$ through \textit{HiGate}.
A lightweight MLP predicts a token wise gate $\mathbf{g}_t\in\mathbb{R}^{N_g\times1}$, and the residual fusion is formulated as 
$\widetilde{\mathbf{F}}_g=\operatorname{LN}(\mathbf{F}_g+\mathbf{g}_t\odot\operatorname{LN}(\mathbf{P}_t))$.
Only the target visual positions are updated by this residual injection, and the resulting features continue through the \emph{Pixel Diffusion Head}.
In this manner, the residual pathway supplies fine-grained spatial cues for image synthesis, while high-level semantic modeling remains in the unified ViT and shared backbone.

\subsection{Task Formulation and Joint Training}
\label{sec:training}

\paragraph{Unified Task Formulation.}
\hiuni{} supports text-to-image generation and image editing through the same generation pathway, with the two tasks differing only in their conditioning context.
For image generation, the condition is a textual prompt $\mathbf{p}$.
For image editing, the condition consists of an editing instruction $\mathbf{p}$ and a reference image $\mathbf{x}_{r}$.
The reference image follows the same visual encoding pathway as the target image and is processed together with the noisy target image by the shared backbone.
Formally, we denote the conditioning context by $\mathbf{C}_{\tau}$, where $\tau\in\{\mathrm{t2i},\mathrm{edit}\}$, $\mathbf{C}_{\mathrm{t2i}}=\mathbf{p}$, and $\mathbf{C}_{\mathrm{edit}}=(\mathbf{p},\mathbf{x}_{r})$.
Given $\mathbf{C}_{\tau}$ and the noisy target image $\mathbf{x}_t$, both tasks follow the same generation pathway without introducing task specific branches.

\paragraph{Joint Training Objectives.}
\hiuni{} jointly optimizes language modeling, image understanding and image generation within the same backbone.
Textual outputs for language modeling and image understanding are trained with the standard next token prediction objective.
Let $\mathbf{C}$ denote the input context, the language modeling objective is $\mathcal{L}_{\mathrm{txt}}=-\frac{1}{M}\sum_{i=1}^{M}\log p_{\theta}(y_i |\mathbf{C},y_{<i})$.
For image generation, we follow JiT~\citep{li2026back} and adopt x-prediction with a velocity-based objective.
Given a clean target image $\mathbf{x}_0$, Gaussian noise $\mathbf{z}\sim\mathcal{N}(0,I)$, and a timestep $t$ sampled such that $\operatorname{logit}(t)\sim\mathcal{N}(-0.8,0.8^2)$, we define $\boldsymbol{\epsilon}_{N_g}=\sigma_{N_g}\mathbf{z}$, where $\sigma_{N_g}$ is a resolution dependent noise scale detailed in Appendix~\ref{app:training_details}.
The noisy image is constructed as $\mathbf{x}_t=t\mathbf{x}_0+(1-t)\boldsymbol{\epsilon}_{N_g}$.
The model takes $\mathbf{x}_t$, timestep $t$, and task condition $\mathbf{C}_{\tau}$ as input and directly predicts the clean image as $\hat{\mathbf{x}}_0=f_{\theta}(\mathbf{x}_t,t,\mathbf{C}_{\tau})$.
The target velocity is $\mathbf{v}=\mathbf{x}_0-\boldsymbol{\epsilon}_{N_g}$.
For numerical stability near $t=1$, the predicted velocity is computed as $\hat{\mathbf{v}}=(\hat{\mathbf{x}}_0-\mathbf{x}_t)/\max(1-t,0.05)$.
The generation objective is then defined as $\mathcal{L}_{\mathrm{gen}}=\mathbb{E}[\|\hat{\mathbf{v}}-\mathbf{v}\|_2^2]$.
The two objectives are jointly optimized as $\mathcal{L}=\mathcal{L}_{\mathrm{txt}}+\lambda_{\mathrm{gen}}\mathcal{L}_{\mathrm{gen}}$, where $\lambda_{\mathrm{gen}}$ balances the two objectives.

\paragraph{Training stages.}
\hiuni{} is trained in three stages consisting of generation warmup, joint pretraining, and supervised fine-tuning.
During warmup, we keep the pretrained backbone and visual encoder frozen, and optimize only the newly initialized \emph{Pixel Diffusion Head} and \emph{HiGate}.
This stage first establishes the mapping from pretrained VLM representations to pixel outputs without perturbing the pretrained components.
Next, we unfreeze the entire model and jointly train text modeling, image understanding, and image generation.
The image resolution is progressively increased during this stage, allowing generation to gradually adapt from low to high resolutions.
Finally, supervised fine-tuning introduces image editing while retaining the existing tasks.

\section{Experiments}
\label{sec:experiments}

We evaluate \hiuni{} on image understanding, image generation, and image editing, and further examine architectural design through controlled comparisons.
Additional training and evaluation details, analyses of pixel diffusion optimization stability and cross task interaction, an extension to chain of thought image generation, and qualitative results are provided in Appendices~\ref{app:training_details}-\ref{app:qualitative_results}.

\subsection{Experimental Setup}
\label{sec:experimental_setup}


\begin{figure}
    \centering
    \includegraphics[width=0.95\linewidth]{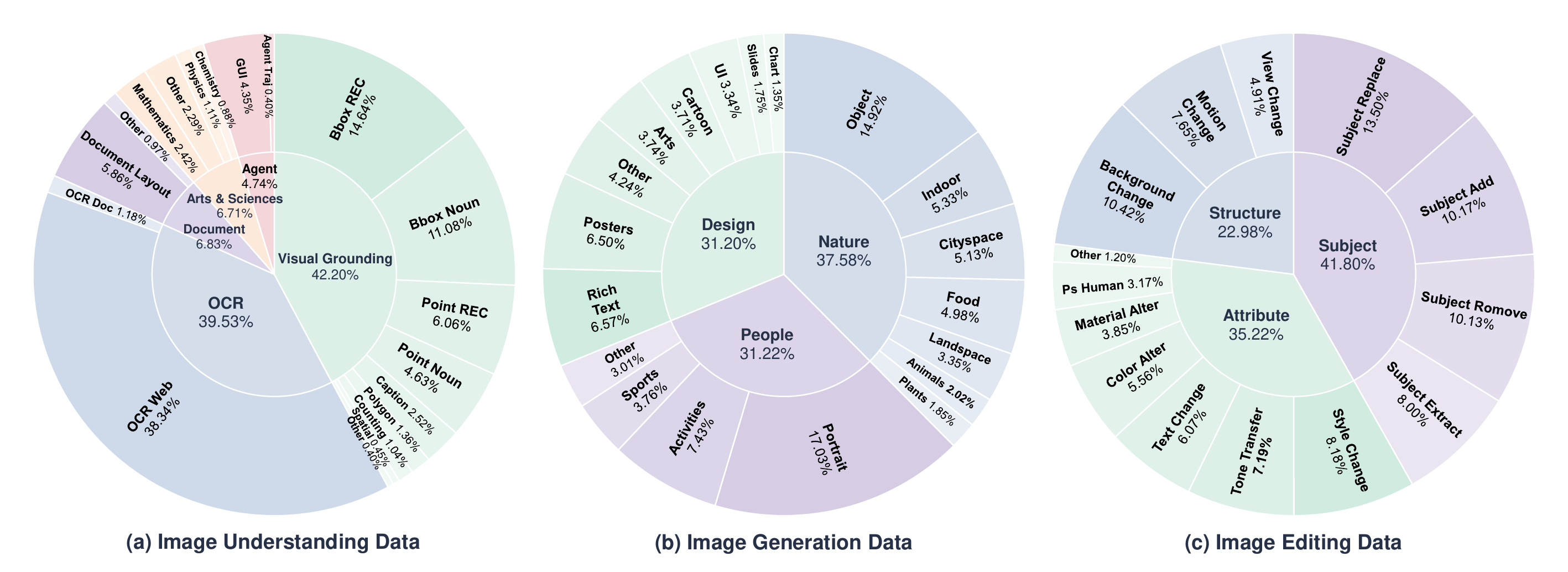}
    \caption{Data distribution of \hiuni's training corpus for image understanding, generation, and editing. Inner rings show the major categories, while outer rings provide finer-grained breakdown.}
    \label{fig:data_distribution}
\end{figure}

\paragraph{Implementation Details.}
We initialize \hiuni{} from Qwen3-VL-4B-Instruct~\citep{bai2025qwen3vl}, reusing its language model backbone as the shared dense backbone and its language aligned ViT as the shared visual encoder.
The multimodal training corpus covers image understanding, image generation, and image editing, with their category distributions summarized in Figure~\ref{fig:data_distribution}.
Following the three-stage procedure in Section~\ref{sec:training}, we first warm up the \emph{Pixel Diffusion Head} and \emph{HiGate} for 30K steps on image generation data while freezing all pretrained components.
We then unfreeze the full model and perform 310K steps of joint pretraining on pure text modeling, image understanding, and image generation, with the generation resolution progressively increased from \(256\times256\) to \(2048\times2048\).
Finally, we conduct 22K steps of supervised fine tuning, where image editing is added while the existing tasks are retained.
The learning rate is \(1\times10^{-4}\) during generation warmup and early joint pretraining, and is reduced to \(2\times10^{-5}\) in later pretraining, while supervised fine tuning uses a peak learning rate of \(2\times10^{-5}\).
The complete phase-wise recipe, including training steps, batch sizes, learning rate schedules, image resolutions, loss weights, and task sampling ratios, is provided in Appendix~\ref{app:training_details} and Table~\ref{tab:hiuni_training_recipe}.

\paragraph{Baselines.}
We compare \hiuni{} with a broad set of UMMs and specialized models.
Representative UMM baselines include Harmon~\citep{wu2025harmonizing}, Cheers~\citep{zhang2026cheers}, InternVL-U~\citep{tian2026internvl}, Janus-Pro~\citep{chen2025janus}, Show-o2~\citep{xie2026show}, TUNA~\citep{liu2026tuna}, Bagel~\citep{deng2025emerging}, SenseNova-U1~\citep{diao2026sensenova}, UniSpace~\citep{yan2026unispace} among others.
For image understanding, we also report Qwen3-VL-4B-Instruct~\citep{bai2025qwen3vl}, from which \hiuni{} is initialized, as a reference for the image understanding capability of the pretrained VLM.
For image generation, we further compare against strong baselines such as GLM-Image~\citep{glmimage2026}, Qwen-Image~\citep{wu2025qwen}, and HiDream-O1~\citep{cai2026hidream}.
For image editing, we include recent specialized models Qwen-Image-Edit-2511~\citep{qwenimageedit2511}, LongCat-Image-Edit~\citep{longcatimageedit}, and Qwen-Image-2.1~\citep{qwenimage21}.
As many earlier benchmarks are nearly saturated, we focus on newer and more discriminative benchmarks here, so the compared models may vary across tables.

\subsection{Main Results}
\label{sec:main_results}

\paragraph{Image Understanding.} 

We evaluate \hiuni{} on eight representative benchmarks, including MathVista~\citep{lu2024mathvista}, MMBench~\citep{liu2024mmbench}, RealWorldQA~\citep{ai2024grok}, MMStar~\citep{chen2024we}, HallusionBench~\citep{guan2024hallusionbench}, AI2D~\citep{kembhavi2016diagram}, OCRBench~\citep{liu2024ocrbench}, and OmniDocBench~\citep{ouyang2025omnidocbench}.
As shown in Table~\ref{tab:understanding_main}, \hiuni{} achieves strong performance across the evaluated understanding benchmarks.
Across the seven higher-is-better benchmarks, \hiuni{} achieves an average score of $75.56$, close to the officially reported $75.61$ of Qwen3-VL-4B-Instruct, while maintaining the same OmniDocBench score of $0.24$.
Compared with the re-evaluated Qwen3-VL-4B-Thinking baseline under the same evaluation settings, \hiuni{} achieves better results on four of eight benchmarks.
Performance changes are benchmark dependent, with improvements on benchmarks such as MathVista and HallusionBench and declines on others such as MMBench and OCRBench.
Despite its relatively compact model scale, \hiuni{} achieves understanding performance comparable to several substantially larger UMMs across the evaluated benchmarks.
Overall, these results indicate that strong understanding performance can coexist with image generation in the proposed shared architecture.

\begin{table*}[t]
    \centering
    \caption{
    Comparison with representative models on image understanding. 
    \hiuni{} is initialized from Qwen3-VL$_{\mathrm{INS}}$-4B and trained in thinking mode.
    We also report official results for Qwen3-VL$_{\mathrm{THK}}$ (marked with $^\star$), alongside our re-evaluated results (marked with $^\dagger$) under identical evaluation settings.
    Model size is reported as understanding backbone parameters + generation-specific parameters.
    Hereinafter the same.}
    \label{tab:understanding_main}
    \scriptsize
    \resizebox{\textwidth}{!}{%
    \begin{tabular}{l c | *{8}{c}}
        \toprule

        \textbf{Models}
        & \textbf{Size}
        & \textbf{MathVista$_{\mathrm{Mini}}$}
        & \textbf{MMBench$_{\mathrm{EN}}$}
        & \textbf{RealWorldQA}
        & \textbf{MMStar}
        & \textbf{HallusionBench}
        & \textbf{AI2D}
        & \textbf{OCRBench}
        & \textbf{OmniDocBench $\downarrow$}
        \\

        \midrule

        \rowcolor{gray!15}
        \multicolumn{10}{c}{\textit{Understanding-specific Models}}
        \\

        Qwen3-VL$_{\mathrm{INS}}^{\star}$
        & 4B
        & 73.7 & 85.1 & 70.9 & 69.8
        & 57.6 & 84.1 & 88.1 & 0.24
        \\

        Qwen3-VL$_{\mathrm{THK}}^{\star}$
        & 4B
        & 79.5 & 86.7 & 73.2 & 73.2
        & 64.1 & 84.9 & 80.8 & 0.23
        \\

        Qwen3-VL$_{\mathrm{THK}}^{\dagger}$
        & 4B
        & 76.5 & 86.2 & 74.1 & 70.3
        & 65.8 & 84.5 & 79.6 & 0.22
        \\

        \midrule

        \rowcolor{gray!15}
        \multicolumn{10}{c}{\textit{Unified Multimodal Models}}
        \\

        Harmon$^{\star}$
        & 1.5B
        & -- & 65.5 & 49.8 & 35.3
        & -- & -- & 11.2 & --
        \\

        Cheers$^{\star}$
        & 1.5B
        & 50.5 & 70.4 & 60.9 & 50.9
        & -- & -- & 58.4 & --
        \\

        InternVL-U
        & 2B + 1.7B
        & 53.9 & 71.7 & 56.9 & 53.5
        & 32.4 & 76.8 & 83.9$^{\star}$ & 0.83
        \\

        Janus-Pro
        & 7B
        & 44.7 & 79.2$^{\star}$ & 58.0 & 48.3
        & 36.4 & 71.3 & 59.0 & 0.81
        \\

        Show-o2
        & 7B
        & 51.8 & 79.3$^{\star}$ & 64.7$^{\star}$ & 56.6
        & 36.2 & 78.6$^{\star}$ & 32.4 & 0.84
        \\

        TUNA$^{\star}$
        & 7B
        & -- & -- & 66.1 & 61.2
        & -- & 79.3 & 74.3 & --
        \\

        Bagel
        & 7B + 7B
        & 73.1$^{\star}$ & 85.0$^{\star}$ & 74.4 & 67.6
        & 51.8 & 89.2 & 80.8 & 0.43
        \\

        SenseNova-U1
        & 8B + 8B
        & 84.2$^{\star}$ & 90.3$^{\star}$ & 73.6$^{\star}$ & 78.3$^{\star}$
        & 67.8$^{\star}$ & 91.7$^{\star}$ & 82.1$^{\star}$ & 0.28
        \\

        \midrule

        \rowcolor{blue!8}
        \textbf{\hiuni}
        & 4B + 0.9B
        & 79.0 & 80.7 & 72.2 & 70.5
        & 66.0 & 86.0 & 74.5 & 0.24
        \\

        \bottomrule
    \end{tabular}%
    }
\end{table*}

\paragraph{Image Generation.} 


We evaluate \hiuni{} on Qwen-Image-Bench~\citep{li2026qwen}, OneIG~\citep{chang2026oneig}, and LongText~\citep{geng2025x}.
These benchmarks provide complementary evaluations that cover perceptual quality, aesthetics, semantic alignment, realism, creativity, and the ability to render long text within images.
From Table~\ref{tab:generation_results}, we can observe that \hiuni{} achieves the highest Chinese score among UMMs on Qwen-Image-Bench, while remaining competitive on the English split.
For OneIG and LongText, \hiuni{} also stays close to the strongest unified baselines, indicating strong performance on both general image generation and long text rendering.
Together, these results show that \hiuni{} achieves strong generation capability within a UMM architecture.

\begin{table*}[t]
    \centering
    \caption{Comparison with representative models on image generation. 
    More results of OneIG are provided in Table~\ref{tab:oneig_results}.
    \textbf{Bold} denotes the best result among UMMs, while \underline{underline} denotes the second-best result. Hereinafter the same.
    }    
    \label{tab:generation_results}

    \scriptsize
    \resizebox{\textwidth}{!}{%
    \begin{tabular}{
        l
        c |
        *{6}{c} | *{6}{c} | *{2}{c} | *{2}{c}
    }
        \toprule
        \multicolumn{1}{l}{\multirow{3}{*}{\textbf{Models}}}
        & \multicolumn{1}{c|}{\multirow{3}{*}{\textbf{Size}}}
        & \multicolumn{6}{c|}{\textbf{Qwen-Image-Bench}$_\mathrm{ZH}$}
        & \multicolumn{6}{c|}{\textbf{Qwen-Image-Bench}$_\mathrm{EN}$}
        & \multicolumn{2}{c|}{\textbf{OneIG}}
        & \multicolumn{2}{c}{\textbf{LongText}}
        \\

        \cmidrule(lr){3-8}
        \cmidrule(lr){9-14}
        \cmidrule(lr){15-16}
        \cmidrule(lr){17-18}

        &
        & Qual.
        & Aesth.
        & Align.
        & Real.
        & Creat.
        & Overall
        & Qual.
        & Aesth.
        & Align.
        & Real.
        & Creat.
        & Overall
        & ZH
        & EN
        & ZH
        & EN
        \\

        \midrule

        \rowcolor{gray!15}
        \multicolumn{18}{c}{
            \textit{Generation-specific Models}
        }
        \\

        GLM-Image
        & 9B + 7B
        & 49.26 & 50.64 & 47.90 & 44.69 & 45.23 & 48.19
        & 49.86 & 49.98 & 47.49 & 44.25 & 44.67 & 47.86
        & 0.511$^{\star}$ & 0.528$^{\star}$
        & 0.979$^{\star}$ & 0.952$^{\star}$
        \\

        Qwen-Image$^\star$
        & 20B
        & 48.44 & 52.25 & 50.72 & 43.16 & 47.30 & 49.23
        & 48.45 & 51.18 & 50.04 & 43.45 & 45.37 & 48.48
        & 0.548 & 0.539
        & 0.946 & 0.943
        \\

        
        HiDream-O1
        & 8B
        & 48.61 & 49.75 & 49.53 & 43.97 & 40.96 & 47.58
        & 48.84 & 49.69 & 49.53 & 43.80 & 41.05 & 47.59
        & 0.518 & 0.521
        & 0.978$^\star$ & 0.979$^\star$
        \\
        
        
        
        \midrule

        \rowcolor{gray!15}
        \multicolumn{18}{c}{
            \textit{Unified Multimodal Models}
        }
        \\

        InternVL-U
        & 2B + 1.7B
        & 42.34 & 43.38 & 40.30 & 38.13 & 34.36 & 40.54
        & 43.92 & 43.03 & 41.09 & 38.87 & 32.35 & 40.82
        & 0.500$^\star$ & 0.500$^\star$
        & 0.860$^\star$ & 0.738$^\star$
        \\

        Bagel
        & 7B + 7B
        & 46.11 & 45.88 & 40.27 & 39.74 & 32.69 & 42.15
        & 45.90 & 47.20 & 43.48 & 40.30 & 34.83 & 43.45
        & 0.370 & 0.361
        & 0.310 & 0.373
        \\

        SenseNova-U1$^\star$
        & 8B + 8B
        & 45.79 & 47.07 & 47.78 & 42.53 & 43.88 & 45.99
        & 47.30 & 50.30 & 50.58 & 43.13 & 46.15 & \textbf{48.28}
        & \textbf{0.535} & 0.549
        & \underline{0.962} & \textbf{0.979}
        \\

        UniSpace
        & 8B + 8B
        & 47.22 & 45.02 & 48.54 & 46.19 & 40.49 & \underline{46.05}
        & 47.44 & 46.54 & 48.94 & 47.54 & 43.84 & 47.05
        & 0.533 $^\star$ & \textbf{0.561}$^\star$
        & -- & --
        \\

        \midrule

        \rowcolor{blue!8}
        \textbf{\hiuni}
        & 4B + 0.9B
        & 46.73 & 50.00 & 50.45 & 43.09 & 45.62 & \textbf{47.98}
        & 47.63 & 50.52 & 48.78 & 43.70 & 44.15 & \underline{47.70}
        & 0.524 & 0.538
        & \textbf{0.963} & \underline{0.954}
        \\

        \bottomrule
    \end{tabular}%
    }
\end{table*}

\begin{table*}[t]
    \centering
    \caption{
    Comparison with representative models on image editing.
    Both benchmarks are judged by GPT-5.2.
    More results of GEdit are provided in Table~\ref{tab:gedit_detail}.}
    \label{tab:editing_detail}

    \newcommand{\hcell}[1]{{\fontsize{5.1}{5.8}\selectfont #1}}
    \scriptsize
    \resizebox{\textwidth}{!}{%
    \begin{tabular}{l c | *{5}{w{c}{2.41em}} w{c}{3.56em} *{4}{w{c}{2.41em}} | *{3}{w{c}{2.41em}} | *{3}{w{c}{2.41em}}}
        \toprule

        \multicolumn{1}{l}{\multirow{2}{*}{\textbf{Models}}}
        & \multicolumn{1}{c|}{\multirow{2}{*}{\textbf{Size}}}
        & \multicolumn{10}{c|}{\textbf{ImgEdit}}
        & \multicolumn{3}{c|}{\textbf{GEdit}$_\mathrm{CN}$}
        & \multicolumn{3}{c}{\textbf{GEdit}$_\mathrm{EN}$}
        \\

        \cmidrule(lr){3-12}
        \cmidrule(lr){13-15}
        \cmidrule(lr){16-18}

        &
        & \hcell{Add} & \hcell{Adjust} & \hcell{Extract} & \hcell{Replace} & \hcell{Remove}
        & \hcell{Background} & \hcell{Style} & \hcell{Hybrid} & \hcell{Action} & \hcell{Overall}
        & \hcell{G\_SC} & \hcell{G\_PQ} & \hcell{G\_O}
        & \hcell{G\_SC} & \hcell{G\_PQ} & \hcell{G\_O}
        \\

        \midrule

        \rowcolor{gray!15}
        \multicolumn{18}{c}{\textit{Editing-specific Models}}
        \\

         LongCat-Image-Edit
        & 6B
        & 4.15 & 4.24 & 3.39 & 4.47 & 3.95 & 3.57 & 4.28 & 3.39 & 4.70 & 4.02
        & 8.13 & 6.44 & 7.15
        & 8.05 & 6.44 & 7.10
        \\

        Qwen-Image-Edit-2511
        & 20B
        & 4.22 & 4.24 & 3.34 & 4.37 & 3.98 & 3.61 & 4.39 & 3.25 & 4.69 & 4.01
        & 8.25 & 6.42 & 7.19
        & 8.20 & 6.46 & 7.17
        \\

        Qwen-Image-2.1
        & 7B
        & 4.17 & 4.24 & 2.16 & 4.48 & 4.13 & 3.75 & 4.27 & 3.15 & 2.89 & 3.69
        & 7.60 & 6.45 & 6.73
        & 7.91 & 6.40 & 6.92
        \\

        \midrule

        \rowcolor{gray!15}
        \multicolumn{18}{c}{\textit{Unified Multimodal Models}}
        \\

        InternVL-U
        & 2B + 1.7B
        & 3.94 & 3.71 & 2.42 & 4.08 & 3.34 & 3.45 & 4.03 & 2.28 & 4.07 & 3.48
        & 6.02 & 5.81 & 5.74
        & 6.05 & 5.80 & 5.74
        \\

        Bagel
        & 7B + 7B
        & 3.49 & 3.36 & 1.17 & 3.69 & 2.82 & 3.06 & 3.80 & 2.59 & 3.82 & 3.09
        & 7.37 & 5.71 & 6.31
        & 7.31 & 5.68 & 6.27
        \\

        SenseNova-U1
        & 8B + 8B
        & 3.82 & 3.96 & 2.44 & 4.20 & 3.11 & 3.59 & 4.47 & 3.09 & 4.20 & 3.65
        & 8.19 & 6.20 & \textbf{7.05}
        & 8.04 & 6.14 & \textbf{6.94}
        \\

        UniSpace
        & 8B + 8B
        & 4.06 & 4.01 & 3.00 & 4.32 & 3.99 & 3.52 & 4.07 & 3.16 & 4.25 & \textbf{3.82}
        & 7.72 & 5.84 & 6.57
        & 7.88 & 5.87 & 6.68
        \\

        \midrule

        \rowcolor{blue!8}
        \textbf{\hiuni}
        & 4B + 0.9B
        & 3.93 & 3.95 & 2.72 & 4.05 & 3.47 & 3.32 & 4.34 & 3.22 & 3.95 & \underline{3.66}
        & 7.87 & 5.88 & \underline{6.70}
        & 7.93 & 6.00 & \underline{6.81}
        \\

        \bottomrule
    \end{tabular}%
    }
\end{table*}

\paragraph{Image Editing} 

For ImgEdit~\citep{ye2026imgedit}, we use GPT-5.2 as the judge model and report scores on a scale $0$ to $5$.
For GEdit~\citep{liu2025step1x}, we follow the VIEScore protocol~\citep{ku2024viescore} but replace the judge backbone with GPT-5.2, which scores semantic consistency (G\_SC) and perceptual quality (G\_PQ) on a scale $0$ to $10$.
As shown in Table~\ref{tab:editing_detail}, \hiuni{} reaches an Overall score of $3.66$ on ImgEdit, ranking second among the compared unified models behind UniSpace ($3.82$) and slightly ahead of SenseNova-U1 ($3.65$), despite its relatively compact model scale.
Editing-only models that are trained exclusively for this task remain ahead, reaching $4.02$ and $4.01$ with LongCat-Image-Edit and Qwen-Image-Edit-2511, respectively.
On GEdit, \hiuni{} obtains overall scores of $6.70$ and $6.81$ on Chinese and English prompts, second among the compared unified models behind SenseNova-U1 ($7.05$ and $6.94$).
These results further indicate the effectiveness of our unified architectural design across diverse and complex tasks.

\subsection{Trade-offs between Shared and Task-Specific Computation}
\label{sec:dense_mot_analysis}

\paragraph{Experimental Configurations.}
In this section, we examine whether understanding and generation can be effectively supported by a shared dense backbone.
We compare three configurations, \hiuni-Dense-2B, \hiuni-MoT-2B, and \hiuni-Dense-4B.
\hiuni-Dense-2B and \hiuni-MoT-2B are both initialized from Qwen3-VL-2B.
For \hiuni-MoT-2B, the pretrained backbone is duplicated to initialize separate understanding and generation branches.
These two models follow the same training recipe and differ only in whether the backbone is shared or task-routed, which both undergo 30K warmup steps followed by 30K pretraining steps.
Moreover, \hiuni-Dense-4B is initialized from Qwen3-VL-4B and is included as an additional reference for allocating a similar total parameter budget to a larger shared backbone. 
We further introduce two single-task baselines, \hiuni-Und-4B and \hiuni-Gen-4B, which share the same architecture and Qwen3-VL-4B initialization as \hiuni-Dense-4B, but are trained exclusively on understanding and generation data, respectively, with the same training steps. 
For a fair comparison, we adopt the same 0.5B diffusion head across all configurations, with its hidden dimension aligned to the 2B backbone.

\paragraph{Results and Analysis.}

As shown in Table~\ref{tab:dense_mot_comparison}, 
compared with \hiuni-MoT-2B, \hiuni-Dense-2B achieves comparable understanding performance ($58.55$ vs. $59.23$) with a fully shared backbone, while its OneIG score decreases from $0.345$ to $0.291$.
This comparison suggests that task-specific backbone computation benefits image generation under the same  training settings, although a shared dense backbone can also support both capabilities.
Under a similar total parameter budget, \hiuni-Dense-4B outperforms \hiuni-MoT-2B on all six understanding benchmarks, while retaining competitive generation performance. 
Notably, the shared design allows \hiuni-Dense-4B to inherit stronger pretrained representations from Qwen3-VL-4B, whereas \hiuni-MoT-2B has to distribute its capacity across two branches initialized from Qwen3-VL-2B.
Within the dense architecture, scaling from 2B to 4B improves the understanding average by 11.1\% and the OneIG score by 7.9\%, suggesting that a larger shared backbone can benefit both capabilities.

\begin{figure*}[t]
    \centering

    \begin{minipage}[t]{0.59\textwidth}
        \centering
        \includegraphics[
            height=0.4\textwidth
        ]{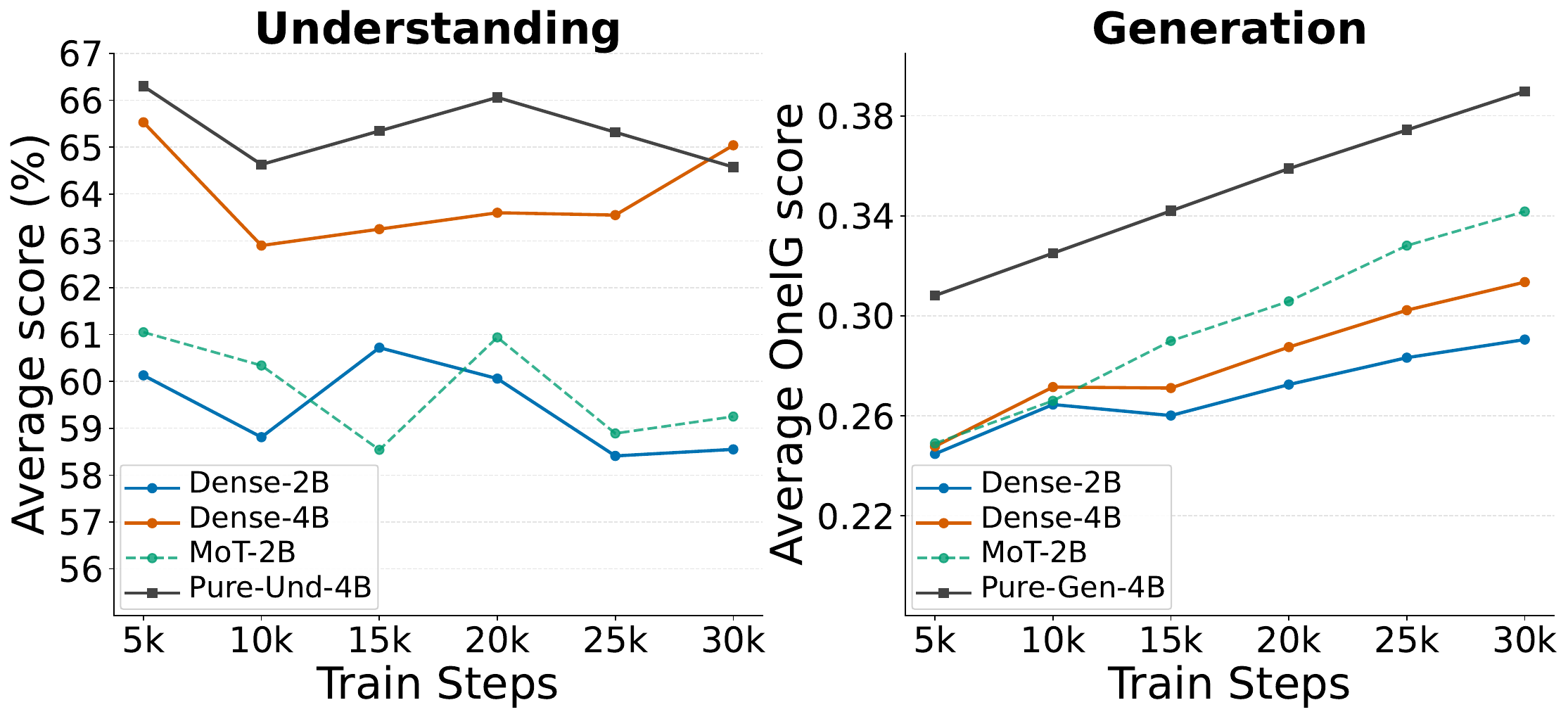}

        \captionof{figure}{
            Training trajectories during the joint pretraining stage after
            warmup. We report the average understanding and OneIG scores here.
        }
        \label{fig:dense_mot_comparison}
    \end{minipage}
    \hfill
    \begin{minipage}[t]{0.39\textwidth}
        \centering
        \includegraphics[
            height=0.6\textwidth,
            trim=0 25 0 0,
            clip
        ]{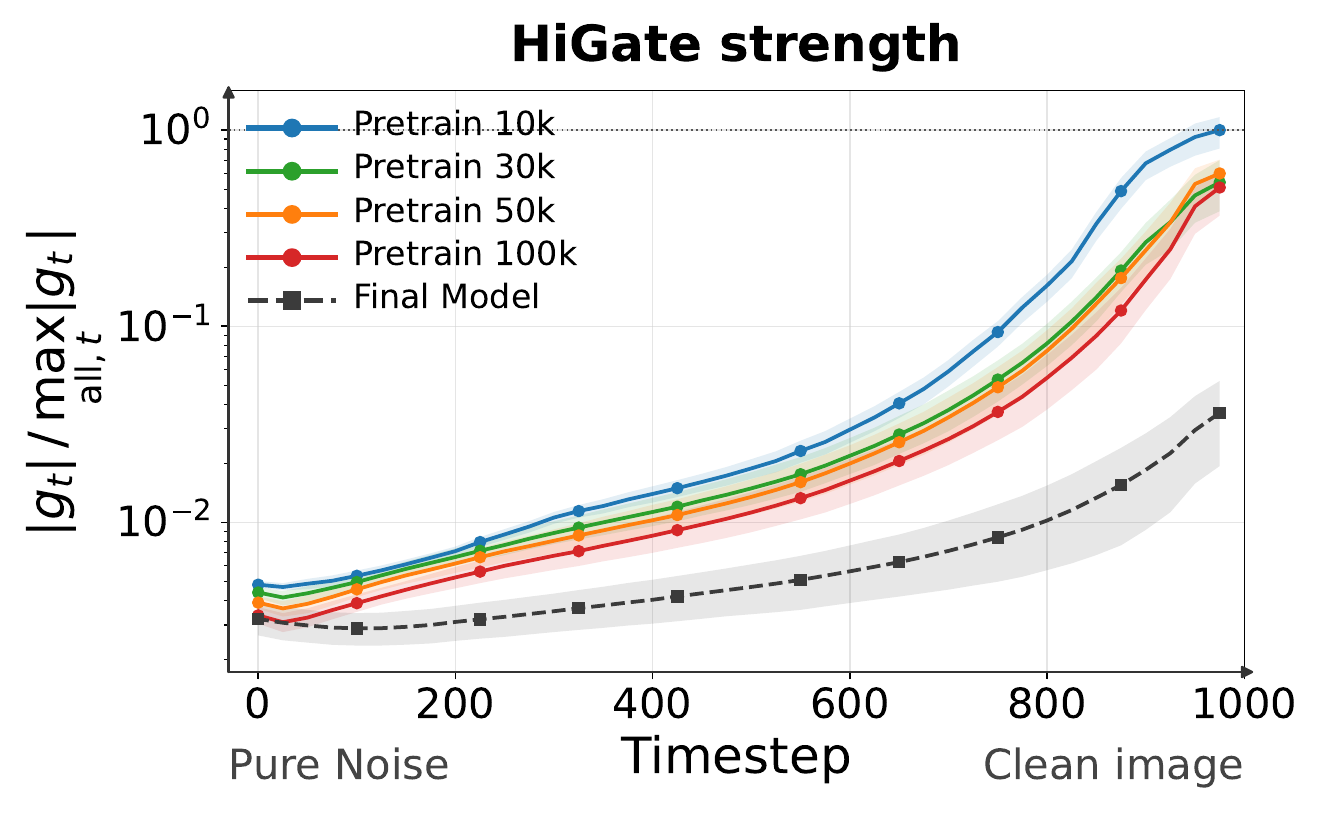}

        \captionof{figure}{Evolution of \emph{HiGate} injection strength across diffusion timesteps and training checkpoints.}
        \label{fig:higate}
    \end{minipage}
\end{figure*}

When comparing with the single-task baselines, \hiuni-Dense-4B achieves comparable understanding performance to \hiuni-Und-4B ($65.04$ vs. $64.58$), suggesting that incorporating image generation does not substantially compromise understanding.
Notably, MMBench remains comparable, while OCRBench shows only a modest decrease, which suggest that the performance gaps relative to the pretrained VLM cannot be solely attributed to joint modeling and may also be influenced by the training data distribution.
Meanwhile, \hiuni-Gen-4B achieves a OneIG score of $0.390$ demonstrating the feasibility of extending a pretrained VLM to image generation by reusing its visual encoder and dense backbone. Although joint training yields a lower generation score of $0.314$, it enables both capabilities within a shared architecture.
Together with the training dynamics in Figure~\ref{fig:dense_mot_comparison}, these results suggest that task-specific backbone computation is not strictly necessary for unified modeling, and further improvements in balancing understanding and generation remain possible.

\begin{table*}[t]
    \centering
    \caption{
        Comparison between dense and task-routed MoT backbones under controlled settings. \hiuni-Dense-2B and \hiuni-MoT-2B differ only in backbone design while \hiuni-Dense-4B examines a different parameter allocation with stronger pretrained initialization.
        \hiuni-Und-4B and \hiuni-Gen-4B serve as understanding-only and generation-only baselines, respectively.
    }
    \label{tab:dense_mot_comparison}

    \scriptsize
    \resizebox{\textwidth}{!}{%
    \begin{tabular}{l c ccccccc c}
        \toprule
        \multicolumn{2}{c}{\textbf{Model}}
        & \multicolumn{7}{c}{\textbf{Understanding}}
        & \textbf{Generation}
        \\
        \cmidrule(lr){1-2}
        \cmidrule(lr){3-9}
        \cmidrule(lr){10-10}

        Name
        & \textbf{Size}
        & \textbf{MMBench$_{\mathrm{EN}}$}
        & \textbf{RealWorldQA}
        & \textbf{MMStar}
        & \textbf{HallusionBench}
        & \textbf{AI2D}
        & \textbf{OCRBench}
        & \textbf{Avg.}
        & \textbf{OneIG Avg.}
        \\
        \midrule

        \rowcolor{blue!8} \textbf{\hiuni-Dense-4B}
        & 4B+0.5B
        & \textbf{72.60}
        & \textbf{65.49}
        & \textbf{53.93}
        & \textbf{48.96}
        & \textbf{77.95}
        & 71.30
        & \textbf{65.04}
        & 0.314
        \\

        \hiuni-Dense-2B
        & 2B+0.5B
        & 64.94 & 60.39 & 46.67 & 38.70 & 70.47 & 70.10
        & 58.55 & 0.291
        \\

        \midrule

        \hiuni-MoT-2B
        & 4B+0.5B
        & 67.97 & 63.27 & 47.00 & 35.33 & 71.76 & 70.02
        & 59.23 & 0.345
        \\

        \midrule

        \hiuni-Und-4B
        & 4B
        & 72.52 & 63.79 & 53.33 & 46.96 & 77.85
        & \textbf{73.00}
        & 64.58 & --
        \\

        \hiuni-Gen-4B
        & 4B+0.5B
        & -- & -- & -- & -- & -- & --
        & -- & \textbf{0.390}
        \\


        \bottomrule
    \end{tabular}%
    }
\end{table*}

\subsection{Analysis of Residual Visual Pathway}

\paragraph{Residual Feature Source.}
The shared ViT and patch merger provide a common semantic representation for understanding and generation, but they also reduce fine-grained spatial information before the diffusion head.
We therefore examine which feature source is more suitable for the residual pathway used in \hiuni.
Specifically, we compare three configurations: directly using pixel-level features before patch embedding (\textit{Pixel Features}), injecting features from the ViT (\textit{ViT Features}), or even removing the residual pathway (\textit{No Gate}).
All other architectural and training settings are kept fixed.
As shown in Table~\ref{tab:gate_input_ablation}, both residual sources improve image generation over \textit{No Gate}.
The average OneIG score increases from $0.267$ to $0.284$ with ViT features and further to $0.314$ with pixel-level features.
Compared with ViT features, pixel-level features provide an additional gain of $0.030$, suggesting that features closer to the pixel space better complement the semantic representation produced by the shared ViT.
Meanwhile, the average understanding score remains almost unchanged across the three settings, ranging from $65.04$ to $65.17$.
These results support our choice of directly injecting pixel-level information into the generation pathway of \hiuni.

\begin{table*}[t]
    \centering
    \caption{
        Analysis of \emph{HiGate} with different insertion positions.
        Pixel Features and ViT Features use features from different stages of
        the visual pathway as the input to \emph{HiGate}, while No Gate removes the gated residual pathway. Pixel Features corresponds to the default design of \hiuni. 
    }
    \label{tab:gate_input_ablation}

    \scriptsize
    \resizebox{\textwidth}{!}{%
    \begin{tabular}{
        l
        ccccccc
        c
    }
        \toprule

        \multicolumn{1}{l}{\multirow{3}{*}{\textbf{Configuration}}}

        & \multicolumn{7}{c}{\textbf{Understanding}}
        & \multicolumn{1}{c}{\textbf{Generation}}
        \\

        \cmidrule(lr){2-8}
        \cmidrule(lr){9-9}

        & \textbf{MMBench$_{\mathrm{EN}}$}
        & \textbf{RealWorldQA}
        & \textbf{MMStar}
        & \textbf{HallusionBench}
        & \textbf{AI2D}
        & \textbf{OCRBench}
        & \textbf{Avg.}
        & \textbf{OneIG Avg.}
        \\

        \midrule

        Pixel Features
        & 72.60
        & 65.49
        & 53.93
        & 48.96
        & 77.95
        & 71.30
        & 65.04
        & \textbf{0.314}
        \\

        ViT Features
        & 74.61
        & 67.45
        & 53.07
        & 49.05
        & 77.82
        & 69.00
        & 65.17
        & 0.284
        \\

        No Gate
        & 72.91
        & 65.88
        & 54.13
        & 46.70
        & 77.75
        & 73.00
        & 65.06
        & 0.267
        \\

        \bottomrule
    \end{tabular}%
    }
\end{table*}

\paragraph{Residual Injection Strength.}
We further analyze how \hiuni{} uses the pixel-level residual pathway during image generation.
Figure~\ref{fig:higate} visualizes the learned gate strength over diffusion timesteps and training checkpoints.
For each timestep, we compute the maximum gate magnitude and normalize it by the global maximum over all plotted timesteps and checkpoints.
The gate remains weak at high noise levels and gradually becomes stronger as the input approaches the clean image, suggesting that the residual pathway contributes mainly at later denoising stages where fine-grained spatial details are more important.
The overall gate strength also decreases as pretraining progresses, and early checkpoints rely more on the residual pathway.
Combined with the residual feature source analysis above, these results suggest that \hiuni{} can retain semantic modeling in the shared backbone while supplying generation-specific spatial details near the output stage.

\subsection{Cross-Task Interaction}
\label{sec:cross_task_interaction}
Beyond architectural sharing, we further examine whether supervision from one visual task affects the others under the shared backbone. 
We chose image editing, as it requires both visual understanding and generation capabilities. For fair comparison, we vary the data proportion under a fixed training token budget.
Specifically, introducing image editing supervision improves the average understanding score from $62.64$ to $64.49$, while the OneIG average score decreases slightly from 0.538 to 0.528. 
These results suggest that cross-task transfer and interference can coexist within a shared backbone, highlighting both the potential and challenges of joint optimization in highly shared UMMs.
Detailed results are provided in Appendix~\ref{app:editing_supervision}.

\subsection{Discussion}

The above findings motivate us to reflect on the broader goals of unified multimodal modeling. We discuss three perspectives on its current challenges and future opportunities.

\paragraph{From coexistence to synergy.}

We believe multimodal unification can be viewed at three levels. Capability unification supports both tasks in one model. Collaborative unification allows understanding, reasoning, and generation to work together. Synergistic unification seeks mutual improvement through shared knowledge and supervision. \hiuni{} mainly explores the first level, while CoT image generation in Appendix~\ref{app:cot_generation} takes an initial step toward the second. 
The editing experiments in Section~\ref{sec:cross_task_interaction} show potential cross-task benefits, but consistent synergy remains an open challenge.

\paragraph{Toward stronger cross-task transfer.}

Beyond better training recipes, data and representation alignment may play a more important role in positive transfer, particularly for fine-grained visual capabilities. 
Recent studies on reconstruction alignment~\citep{xie2026reconstruction}, multi-representation generation~\citep{su2026generation}, and capability transfer~\citep{kang2026transferability} support this direction.
Structured data representations, such as XML-style captions or even image-related code, may connect semantic understanding and visual generation more directly than conventional descriptions.

\paragraph{Native unification and agentic systems.}

Agentic systems can manage long-term memory, planning, and tool use without native interleaved generation. Native UMMs, however, may better support closely coupled understanding, reasoning, and generation within a shared context. We see these approaches as complementary, with agents handling long-term coordination and UMMs supporting close multimodal interactions. Whether native interaction offers advantages over modular systems remains an important open question.

\section{Conclusion}
\label{sec:conclusion}

We present \hiuni, a unified multimodal model that harmonizes visual understanding and generation within a shared visual space.
\hiuni{} reuses the language aligned ViT and dense Transformer of a pretrained VLM for shared semantic modeling, while supplying generation specific spatial details only near the output stage.
Experiments show that strong understanding and generation can coexist within such a highly shared architecture, without requiring separate visual encoders or task specific backbones.
Beyond \hiuni{} itself, we believe this simple and scalable design provides a promising path to extend mature VLMs into more unified models with richer generative capabilities, broader tasks, and more interactive multimodal scenarios.




\section{Authors}
\label{sec:authors}

Authors are listed in alphabetical order of the first name:
\begin{itemize}
\item \textbf{Project Lead}: Zhenyu Zhang
\item \textbf{Core Contributors}: Jiahao Hu, Jingdong Du, Yichen Zhang, Yi Song
\item \textbf{Contributors}: Jiaxin Niu, Ruihang Li, Xiangzhao Hao, Yiqian Zhang, Yong Huang, Zefeng Zhang, Zheyuan Gu
\item \textbf{Advisor}: Shuohuan Wang
\end{itemize}





\bibliography{iclr2027_conference}

@inproceedings{wu2025janus,
  title={Janus: Decoupling visual encoding for unified multimodal understanding and generation},
  author={Wu, Chengyue and Chen, Xiaokang and Wu, Zhiyu and Ma, Yiyang and Liu, Xingchao and Pan, Zizheng and Liu, Wen and Xie, Zhenda and Yu, Xingkai and Ruan, Chong and others},
  booktitle={2025 IEEE/CVF Conference on Computer Vision and Pattern Recognition (CVPR)},
  pages={12966--12977},
  year={2025},
  organization={IEEE}
}

@inproceedings{ma2025janusflow,
  title={Janusflow: Harmonizing autoregression and rectified flow for unified multimodal understanding and generation},
  author={Ma, Yiyang and Liu, Xingchao and Chen, Xiaokang and Liu, Wen and Wu, Chengyue and Wu, Zhiyu and Pan, Zizheng and Xie, Zhenda and Zhang, Haowei and Yu, Xingkai and others},
  booktitle={2025 IEEE/CVF Conference on Computer Vision and Pattern Recognition (CVPR)},
  pages={7739--7751},
  year={2025},
  organization={IEEE}
}

@article{team2025hunyuanimage,
  title={HunyuanImage 3.0 Technical Report},
  author={Team, Tencent Hunyuan Foundation Model},
  journal={arXiv preprint arXiv:2509.23951},
  year={2025}
}

@article{deng2025emerging,
  title={Emerging properties in unified multimodal pretraining},
  author={Deng, Chaorui and Zhu, Deyao and Li, Kunchang and Gou, Chenhui and Li, Feng and Wang, Zeyu and Zhong, Shu and Yu, Weihao and Nie, Xiaonan and Song, Ziang and others},
  journal={arXiv preprint arXiv:2505.14683},
  year={2025}
}

@article{agarwal2026cosmos,
  title={Cosmos 3: Omnimodal world models for physical ai},
  author={Agarwal, Niket and Ali, Arslan and Allen, Jon and Antolini, Martin and Aubame, Adeline and Azzolini, Alisson and Bai, Junjie and Bala, Maciej and Balaji, Yogesh and Bapst, Josh and others},
  journal={arXiv preprint arXiv:2606.02800},
  year={2026}
}

@inproceedings{xie2026reconstruction,
  title={Reconstruction alignment improves unified multimodal models},
  author={Xie, Ji and Zettlemoyer, Luke and Wang, Xudong and others},
  booktitle={International Conference on Learning Representations},
  volume={2026},
  pages={120095--120137},
  year={2026}
}

@InProceedings{su2026generation,
  title = 	 {Generation Enhances Understanding in Unified Multimodal Models via Multi-Representation Generation},
  author =       {Su, Zihan and Wei, Hongyang and Cen, Kangrui and Wang, Yong and Chen, Guanhua and Yuan, Chun and Chu, Xiangxiang},
  booktitle = 	 {Proceedings of the 43rd International Conference on Machine Learning},
  pages = 	 {116131--116146},
  year = 	 {2026},
  editor = 	 {Zhang, Tong and Dudik, Miroslav and Jaggi, Martin and Agarwal, Alekh and Li, Sharon and Schuurmans, Dale and Zhu, Jerry and Berkenkamp, Felix and Dong, Hanze and Bietti, Alberto},
  volume = 	 {306},
  series = 	 {Proceedings of Machine Learning Research},
  month = 	 {06--11 Jul},
  publisher =    {PMLR},
  url = 	 {https://proceedings.mlr.press/v306/su26n.html}
}

@inproceedings{kang2026transferability,
  title={Transferability between understanding and generation in unified multimodal models},
  author={Kang, Jiwon and Yoon, Heeji and Jung, Jaewoo and Min, Jaewon and Jeon, Minkyeong and Hwang, Biyeon and Jung, Sangwon and Kim, Seungryong},
  booktitle={European Conference on Computer Vision},
  pages={626--647},
  year={2026},
  organization={Springer}
}

@article{yan2026unispace,
  title={UniSpace: Unified Visual Representation and Scalable Multimodal Modeling},
  author={Yan, Jinbo and Qiao, Limeng and Qin, Jie and He, Junyan and Wu, Feize and Wan, Guanglu},
  journal={arXiv preprint arXiv:2608.08676},
  year={2026}
}

@article{wu2026uncovering,
  title={Uncovering Understanding-Generation Synergy in Native Unified Multimodal Models: From Representation, Task to System},
  author={Wu, Penghao and Diao, Haiwen and Fan, Weichen and Lu, Lewei and Lin, Dahua and Liu, Ziwei},
  journal={arXiv preprint arXiv:2609.01607},
  year={2026}
}

@article{diao2026sensenova,
  title={Sensenova-u1: Unifying multimodal understanding and generation with neo-unify architecture},
  author={Diao, Haiwen and Wu, Penghao and Deng, Hanming and Wang, Jiahao and Bai, Shihao and Wu, Silei and Fan, Weichen and Ye, Wenjie and Tong, Wenwen and Fan, Xiangyu and others},
  journal={arXiv preprint arXiv:2605.12500},
  year={2026}
}

@article{zheng2025aia,
  title={AIA: Rethinking Architecture Decoupling Strategy In Unified Multimodal Model},
  author={Zheng, Dian and Zhang, Manyuan and Li, Hongyu and Zou, Kai and Liu, Hongbo and Guo, Ziyu and Feng, Kaituo and Liu, Yexin and Luo, Ying and Li, Hongsheng},
  journal={arXiv preprint arXiv:2511.22663},
  year={2025}
}

@article{zhang2026hydra,
  title={HYDRA-X: Native Unified Multimodal Models with Holistic Visual Tokenizers},
  author={Zhang, Guozhen and Qiu, Xuerui and Cui, Yutao and Song, Tianhui and Li, Changlin and Li, Junzhe and Huang, Tao and Zhang, Xiao and Li, Yang and Wu, Jianbing and others},
  journal={arXiv preprint arXiv:2606.13289},
  year={2026}
}

@inproceedings{liu2026tuna,
  title={Tuna: Taming unified visual representations for native unified multimodal models},
  author={Liu, Zhiheng and Ren, Weiming and Liu, Haozhe and Zhou, Zijian and Chen, Shoufa and Qiu, Haonan and Huang, Xiaoke and An, Zhaochong and Yang, Fanny and Patel, Aditya and others},
  booktitle={Proceedings of the IEEE/CVF Conference on Computer Vision and Pattern Recognition},
  pages={15740--15751},
  year={2026}
}

@article{liu2026tuna2,
  title={Tuna-2: Pixel embeddings beat vision encoders for multimodal understanding and generation},
  author={Liu, Zhiheng and Ren, Weiming and Huang, Xiaoke and Chen, Shoufa and Li, Tianhong and Chen, Mengzhao and Ji, Yatai and He, Sen and Schult, Jonas and Zeng, Belinda and others},
  journal={arXiv preprint arXiv:2604.24763},
  year={2026}
}

@inproceedings{chen2026dip,
  title={DiP: Taming Diffusion Models in Pixel Space},
  author={Chen, Zhennan and Zhu, Junwei and Chen, Xu and Zhang, Jiangning and Hu, Xiaobin and Zhao, Hanzhen and Wang, Chengjie and Yang, Jian and Tai, Ying},
  booktitle={Proceedings of the IEEE/CVF Conference on Computer Vision and Pattern Recognition},
  pages={36136--36146},
  year={2026}
}

@inproceedings{ma2026deco,
  title={DeCo: Frequency-Decoupled Pixel Diffusion for End-to-End Image Generation},
  author={Ma, Zehong and Wei, Longhui and Wang, Shuai and Zhang, Shiliang and Tian, Qi},
  booktitle={Proceedings of the IEEE/CVF Conference on Computer Vision and Pattern Recognition},
  pages={43600--43610},
  year={2026}
}

@article{pan2026repfusion,
  title={RepFusion: Leveraging Multimodal Priors for Denoising in Representation Space},
  author={Pan, Xichen and Singh, Aashu and Shukla, Satya Narayan and Fan, Xiangjun and Mishra, Shlok Kumar and Xie, Saining},
  journal={arXiv preprint arXiv:2606.14700},
  year={2026}
}

@inproceedings{wu2025vila,
  title={Vila-u: a unified foundation model integrating visual understanding and generation},
  author={Wu, Yecheng and Zhang, Zhuoyang and Chen, Junyu and Tang, Haotian and Li, Dacheng and Fang, Yunhao and Zhu, Ligeng and Xie, Enze and Yin, Hongxu and Yi, Li and others},
  booktitle={International Conference on Learning Representations},
  volume={2025},
  pages={93620--93638},
  year={2025}
}

@article{ma2026unitok,
  title={Unitok: A unified tokenizer for visual generation and understanding},
  author={Ma, Chuofan and Jiang, Yi and Wu, Junfeng and Yang, Jihan and Yu, Xin and Yuan, Zehuan and Peng, Bingyue and Qi, Xiaojuan},
  journal={Advances in Neural Information Processing Systems},
  volume={38},
  pages={129274--129297},
  year={2026}
}

@inproceedings{yu2026pixeldit,
  title={Pixeldit: Pixel diffusion transformers for image generation},
  author={Yu, Yongsheng and Xiong, Wei and Nie, Weili and Sheng, Yichen and Liu, Shiqiu and Luo, Jiebo},
  booktitle={Proceedings of the IEEE/CVF Conference on Computer Vision and Pattern Recognition},
  pages={14273--14282},
  year={2026}
}

@inproceedings{li2026back,
  title={Back to basics: Let denoising generative models denoise},
  author={Li, Tianhong and He, Kaiming},
  booktitle={Proceedings of the IEEE/CVF Conference on Computer Vision and Pattern Recognition},
  pages={36115--36125},
  year={2026}
}

@article{zhang2026cheers,
  title={Cheers: Decoupling Patch Details from Semantic Representations Enables Unified Multimodal Comprehension and Generation},
  author={Zhang, Yichen and Peng, Da and Guo, Zonghao and Zhang, Zijian and Yang, Xuesong and Sun, Tong and Sun, Shichu and Zhang, Yidan and Li, Yanghao and Zhao, Haiyan and others},
  journal={arXiv preprint arXiv:2603.12793},
  year={2026}
}

@article{bai2025qwen3vl,
  title={Qwen3-vl technical report},
  author={Bai, Shuai and Cai, Yuxuan and Chen, Ruizhe and Chen, Keqin and Chen, Xionghui and Cheng, Zesen and Deng, Lianghao and Ding, Wei and Gao, Chang and Ge, Chunjiang and others},
  journal={arXiv preprint arXiv:2511.21631},
  year={2025}
}

@article{wang2025internvl35,
  title={Internvl3.5: Advancing open-source multimodal models in versatility, reasoning, and efficiency},
  author={Wang, Weiyun and Gao, Zhangwei and Gu, Lixin and Pu, Hengjun and Cui, Long and Wei, Xingguang and Liu, Zhaoyang and Jing, Linglin and Ye, Shenglong and Shao, Jie and others},
  journal={arXiv preprint arXiv:2508.18265},
  year={2025}
}

@article{team2024chameleon,
  title={Chameleon: Mixed-modal early-fusion foundation models},
  author={Team, Chameleon},
  journal={arXiv preprint arXiv:2405.09818},
  year={2024}
}

@article{cui2025emu35,
  title={Emu3.5: Native multimodal models are world learners},
  author={Cui, Yufeng and Chen, Honghao and Deng, Haoge and Huang, Xu and Li, Xinghang and Liu, Jirong and Liu, Yang and Luo, Zhuoyan and Wang, Jinsheng and Wang, Wenxuan and others},
  journal={arXiv preprint arXiv:2510.26583},
  year={2025}
}

@article{team2026longcatnext,
  title={Longcat-next: Lexicalizing modalities as discrete tokens},
  author={Team, Meituan LongCat and Xiao, Bin and Wang, Chao and Li, Chengjiang and Zhang, Chi and Peng, Chong and Yu, Hang and Yang, Hao and Yan, Haonan and Sun, Haoze and others},
  journal={arXiv preprint arXiv:2603.27538},
  year={2026}
}

@article{wang2026ernie5,
  title={Ernie 5.0 technical report},
  author={Wang, Haifeng and Wu, Hua and Wu, Tian and Sun, Yu and Liu, Jing and Yu, Dianhai and Ma, Yanjun and He, Jingzhou and He, Zhongjun and Hong, Dou and others},
  journal={arXiv preprint arXiv:2602.04705},
  year={2026}
}

@inproceedings{zhou2025transfusion,
  title={Transfusion: Predict the next token and diffuse images with one multi-modal model},
  author={Zhou, Chunting and Yu, Lili and Babu, Arun and Tirumala, Kushal and Yasunaga, Michihiro and Shamis, Leonid and Kahn, Jacob and Ma, Xuezhe and Zettlemoyer, Luke and Levy, Omer},
  booktitle={International Conference on Learning Representations},
  volume={2025},
  pages={6446--6469},
  year={2025}
}

@article{jiang2026empirical,
  title={An Empirical Study of Training Pixel-Space Text-to-Image Diffusion Models},
  author={Jiang, Dengyang and Du, Ruoyi and Chen, Zhennan and Liu, Dongyang and Wang, Zanyi and Zheng, Mingzhe and Yang, Xiangpeng and Cai, Huanqia and Hao, Aiming and Jiang, Yuming and others},
  journal={arXiv preprint arXiv:2608.16887},
  year={2026}
}

@inproceedings{peebles2023scalable,
  title={Scalable diffusion models with transformers},
  author={Peebles, William and Xie, Saining},
  booktitle={2023 IEEE/CVF International Conference on Computer Vision (ICCV)},
  pages={4172--4182},
  year={2023},
  organization={IEEE}
}

@article{xie2026show,
  title={Show-o2: Improved native unified multimodal models},
  author={Xie, Jinheng and Yang, Zhenheng and Shou, Mike Zheng},
  journal={Advances in Neural Information Processing Systems},
  volume={38},
  pages={47490--47518},
  year={2026}
}

@article{tian2026internvl,
  title={Internvl-u: Democratizing unified multimodal models for understanding, reasoning, generation and editing},
  author={Tian, Changyao and Yang, Danni and Chen, Guanzhou and Cui, Erfei and Wang, Zhaokai and Duan, Yuchen and Yin, Penghao and Chen, Sitao and Yang, Ganlin and Liu, Mingxin and others},
  journal={arXiv preprint arXiv:2603.09877},
  year={2026}
}

@article{chen2025janus,
  title={Janus-pro: Unified multimodal understanding and generation with data and model scaling},
  author={Chen, Xiaokang and Wu, Zhiyu and Liu, Xingchao and Pan, Zizheng and Liu, Wen and Xie, Zhenda and Yu, Xingkai and Ruan, Chong},
  journal={arXiv preprint arXiv:2501.17811},
  year={2025}
}

@inproceedings{wu2025harmonizing,
  title={Harmonizing visual representations for unified multimodal understanding and generation},
  author={Wu, Size and Zhang, Wenwei and Xu, Lumin and Jin, Sheng and Wu, Zhonghua and Tao, Qingyi and Liu, Wentao and Li, Wei and Loy, Chen Change},
  booktitle={2025 IEEE/CVF International Conference on Computer Vision (ICCV)},
  pages={17739--17750},
  year={2025},
  organization={IEEE}
}

@inproceedings{lu2024mathvista,
  title={Mathvista: Evaluating mathematical reasoning of foundation models in visual contexts},
  author={Lu, Pan and Bansal, Hritik and Xia, Tony and Liu, Jiacheng and Li, Chunyuan and Hajishirzi, Hannaneh and Cheng, Hao and Chang, Kai-Wei and Galley, Michel and Gao, Jianfeng},
  booktitle={International Conference on Learning Representations},
  volume={2024},
  pages={23439--23554},
  year={2024}
}

@inproceedings{liu2024mmbench,
  title={Mmbench: Is your multi-modal model an all-around player?},
  author={Liu, Yuan and Duan, Haodong and Zhang, Yuanhan and Li, Bo and Zhang, Songyang and Zhao, Wangbo and Yuan, Yike and Wang, Jiaqi and He, Conghui and Liu, Ziwei and others},
  booktitle={European conference on computer vision},
  pages={216--233},
  year={2024},
  organization={Springer}
}

@misc{ai2024grok,
  title={Grok-1.5 vision preview},
  author={AI, X},
  url={https://x.ai/blog/grok-1.5v},
  year={2024}
}

@article{chen2024we,
  title={Are we on the right way for evaluating large vision-language models?},
  author={Chen, Lin and Li, Jinsong and Dong, Xiaoyi and Zhang, Pan and Zang, Yuhang and Chen, Zehui and Duan, Haodong and Wang, Jiaqi and Qiao, Yu and Lin, Dahua and others},
  journal={Advances in Neural Information Processing Systems},
  volume={37},
  pages={27056--27087},
  year={2024}
}

@inproceedings{guan2024hallusionbench,
  title={Hallusionbench: an advanced diagnostic suite for entangled language hallucination and visual illusion in large vision-language models},
  author={Guan, Tianrui and Liu, Fuxiao and Wu, Xiyang and Xian, Ruiqi and Li, Zongxia and Liu, Xiaoyu and Wang, Xijun and Chen, Lichang and Huang, Furong and Yacoob, Yaser and others},
  booktitle={2024 IEEE/CVF Conference on Computer Vision and Pattern Recognition (CVPR)},
  pages={14375--14385},
  year={2024},
  organization={IEEE}
}

@inproceedings{kembhavi2016diagram,
  title={A diagram is worth a dozen images},
  author={Kembhavi, Aniruddha and Salvato, Mike and Kolve, Eric and Seo, Minjoon and Hajishirzi, Hannaneh and Farhadi, Ali},
  booktitle={European conference on computer vision},
  pages={235--251},
  year={2016},
  organization={Springer}
}

@article{liu2024ocrbench,
  title={Ocrbench: on the hidden mystery of ocr in large multimodal models},
  author={Liu, Yuliang and Li, Zhang and Huang, Mingxin and Yang, Biao and Yu, Wenwen and Li, Chunyuan and Yin, Xu-Cheng and Liu, Cheng-Lin and Jin, Lianwen and Bai, Xiang},
  journal={Science China Information Sciences},
  volume={67},
  number={12},
  pages={220102},
  year={2024},
  publisher={Springer}
}

@inproceedings{ouyang2025omnidocbench,
  title={Omnidocbench: Benchmarking diverse pdf document parsing with comprehensive annotations},
  author={Ouyang, Linke and Qu, Yuan and Zhou, Hongbin and Zhu, Jiawei and Zhang, Rui and Lin, Qunshu and Wang, Bin and Zhao, Zhiyuan and Jiang, Man and Zhao, Xiaomeng and others},
  booktitle={2025 IEEE/CVF Conference on Computer Vision and Pattern Recognition (CVPR)},
  pages={24838--24848},
  year={2025},
  organization={IEEE}
}

@misc{glmimage2026,
  title={GLM-Image: Auto-regressive for Dense-knowledge and High-fidelity Image Generation},
  author={{GLM-Image Team}},
  year={2026},
}

@article{cai2026hidream,
  title={Hidream-o1-image: A natively unified image generative foundation model with pixel-level unified transformer},
  author={Cai, Qi and Chen, Jingwen and Gao, Chengmin and Gong, Zijian and Li, Yehao and Pan, Yingwei and Peng, Yi and Qiu, Zhaofan and Yu, Kai and Zhang, Yiheng and others},
  journal={arXiv preprint arXiv:2605.11061},
  year={2026}
}

@article{wu2025qwen,
  title={Qwen-image technical report},
  author={Wu, Chenfei and Li, Jiahao and Zhou, Jingren and Lin, Junyang and Gao, Kaiyuan and Yan, Kun and Yin, Sheng-ming and Bai, Shuai and Xu, Xiao and Chen, Yilei and others},
  journal={arXiv preprint arXiv:2508.02324},
  year={2025}
}

@misc{qwenimage21,
  title={Qwen-Image 2.1},
  author={{Qwen Team}},
  year={2026},
  howpublished={\url{https://github.com/QwenLM/Qwen-Image-2.1}},
}

@article{li2026qwen,
  title={Qwen-image-bench: From generation to creation in text-to-image evaluation},
  author={Li, Niantong and Hu, Guangzheng and Qiao, Weixu and Ba, Ying and Hong, Qichen and Shen, Shijun and Wang, Jinlin and Zhou, Fan and Kang, Jianye and Shang, Xin and others},
  journal={arXiv preprint arXiv:2605.28091},
  year={2026}
}

@article{chang2026oneig,
  title={Oneig-bench: Omni-dimensional nuanced evaluation for image generation},
  author={Chang, Jingjing and Fang, Yixiao and Xing, Peng and Wu, Shuhan and Cheng, Wei and Wang, Rui and Zeng, Xianfang and Yu, Gang and Chen, Hai-Bao},
  journal={Advances in Neural Information Processing Systems},
  volume={38},
  year={2026}
}

@article{geng2025x,
  title={X-omni: Reinforcement learning makes discrete autoregressive image generative models great again},
  author={Geng, Zigang and Wang, Yibing and Ma, Yeyao and Li, Chen and Rao, Yongming and Gu, Shuyang and Zhong, Zhao and Lu, Qinglin and Hu, Han and Zhang, Xiaosong and others},
  journal={arXiv preprint arXiv:2507.22058},
  year={2025}
}

@article{ye2026imgedit,
  title={Imgedit: A unified image editing dataset and benchmark},
  author={Ye, Yang and He, Xianyi and Li, Zongjian and Yuan, Shenghai and Yan, Zhiyuan and Hou, Bohan and Yuan, Li and others},
  journal={Advances in Neural Information Processing Systems},
  volume={38},
  year={2026}
}

@article{liu2025step1x,
  title={Step1x-edit: A practical framework for general image editing},
  author={Liu, Shiyu and Han, Yucheng and Xing, Peng and Yin, Fukun and Wang, Rui and Cheng, Wei and Liao, Jiaqi and Wang, Yingming and Fu, Honghao and Han, Chunrui and others},
  journal={arXiv preprint arXiv:2504.17761},
  year={2025}
}

@inproceedings{wei2025omniedit,
  title={Omniedit: Building image editing generalist models through specialist supervision},
  author={Wei, Cong and Xiong, Zheyang and Ren, Weiming and Du, Xeron and Zhang, Ge and Chen, Wenhu},
  booktitle={International Conference on Learning Representations},
  volume={2025},
  pages={259--271},
  year={2025}
}

@article{bai2024humanedit,
  title={Humanedit: A high-quality human-rewarded dataset for instruction-based image editing},
  author={Bai, Jinbin and Chow, Wei and Yang, Ling and Li, Xiangtai and Li, Juncheng and Zhang, Hanwang and Yan, Shuicheng},
  journal={arXiv preprint arXiv:2412.04280},
  year={2024}
}

@article{chen2026scaleedit,
  title={ScaleEdit-12M: Scaling Open-Source Image Editing Data Generation via Multi-Agent Framework},
  author={Chen, Guanzhou and Cui, Erfei and Tian, Changyao and Yang, Danni and Yang, Ganlin and Qiao, Yu and Li, Hongsheng and Luo, Gen and Zhang, Hongjie},
  journal={arXiv preprint arXiv:2603.20644},
  year={2026}
}

@article{chen2026vins,
  title={VINS-120K: Ultra High-Resolution Image Editing with A Large-Scale Dataset},
  author={Chen, Zhizhou and Guan, Shanyan and Gao, Zhanxin and Ci, En and Ge, Yanhao and Li, Wei and Zhang, Zhenyu and Yang, Jian and Tai, Ying},
  journal={arXiv preprint arXiv:2605.23518},
  year={2026}
}

@article{wu2025omnigen2,
  title={Omnigen2: Exploration to advanced multimodal generation},
  author={Wu, Chenyuan and Zheng, Pengfei and Yan, Ruiran and Xiao, Shitao and Luo, Xin and Wang, Yueze and Li, Wanli and Jiang, Xiyan and Liu, Yexin and Zhou, Junjie and others},
  journal={arXiv e-prints},
  pages={arXiv--2506},
  year={2025}
}

@article{hu2024vivid,
  title={Vivid-10m: A dataset and baseline for versatile and interactive video local editing},
  author={Hu, Jiahao and Zhong, Tianxiong and Wang, Xuebo and Jiang, Boyuan and Tian, Xingye and Yang, Fei and Wan, Pengfei and Zhang, Di},
  journal={arXiv preprint arXiv:2411.15260},
  year={2024}
}

@article{wang2026unireason,
  title={Unireason 1.0: A unified reasoning framework for world knowledge aligned image generation and editing},
  author={Wang, Dianyi and Ma, Chaofan and Han, Feng and Wu, Size and Song, Wei and Wang, Yibin and Zhang, Zhixiong and Wang, Tianhang and Wang, Siyuan and Wei, Zhongyu and others},
  journal={arXiv preprint arXiv:2602.02437},
  year={2026}
}

@article{niu2025wise,
  title={Wise: A world knowledge-informed semantic evaluation for text-to-image generation},
  author={Niu, Yuwei and Ning, Munan and Zheng, Mengren and Jin, Weiyang and Lin, Bin and Jin, Peng and Liao, Jiaqi and Feng, Chaoran and Meng, Fanqing and Ning, Kunpeng and others},
  journal={arXiv preprint arXiv:2503.07265},
  year={2025}
}

@misc{longcatimageedit,
  title={LongCat-Image-Edit},
  author={{Meituan LongCat Team}},
  year={2026},
  howpublished={\url{https://huggingface.co/meituan-longcat/LongCat-Image-Edit}},
}

@misc{qwenimageedit2511,
  title={Qwen-Image-Edit-2511},
  author={{Qwen Team}},
  year={2025},
  howpublished={\url{https://huggingface.co/Qwen/Qwen-Image-Edit-2511}},
}

@inproceedings{ku2024viescore,
  title={Viescore: Towards explainable metrics for conditional image synthesis evaluation},
  author={Ku, Max and Jiang, Dongfu and Wei, Cong and Yue, Xiang and Chen, Wenhu},
  booktitle={Proceedings of the 62nd Annual Meeting of the Association for Computational Linguistics (Volume 1: Long Papers)},
  pages={12268--12290},
  year={2024}
}
\bibliographystyle{iclr2027_conference}

\newpage

\appendix

\section{Training Details}
\label{app:training_details}

\begin{table*}[t]
    \centering

    \caption{
        Training recipe of \hiuni.
        Image resolutions denote the maximum resolution of an image, subject to the image number and sequence length constraint.
        ``-'' denotes an inapplicable value.
    }

    \resizebox{\textwidth}{!}{%
    \begin{tabular}{l|c|c|c|c|c|c|c}
        \toprule

        & \textbf{Generation}
        & \multicolumn{4}{c|}{\textbf{Pretraining}}
        & \multicolumn{2}{c}{\textbf{SFT}}
        \\

        &
        \makecell{\textbf{Warmup}}
        &
        \makecell{Phase I}
        &
        \makecell{Phase II}
        &
        \makecell{Phase III}
        &
        \makecell{Phase IV}
        &
        \makecell{Phase I}
        &
        \makecell{Phase II}
        \\

        \midrule

        \multicolumn{8}{l}{\textbf{Hyperparameters}}
        \\

        Peak learning rate
        & $1 \times 10^{-4}$
        & $1 \times 10^{-4}$
        & $1 \times 10^{-4}$
        & $1 \times 10^{-4}$
        & $2 \times 10^{-5}$
        & $2 \times 10^{-5}$
        & $2 \times 10^{-5}$
        \\

        Min learning rate
        & -
        & -
        & -
        & $2 \times 10^{-5}$
        & -
        & -
        & $2 \times 10^{-6}$
        \\

        LR scheduler
        & Constant
        & Constant
        & Constant
        & Cosine decay
        & Constant
        & Constant
        & Cosine decay
        \\

        Optimizer
        & \multicolumn{7}{c}{
            AdamW $(\beta_1=0.9,\ \beta_2=0.95,\ \epsilon=10^{-8})$
        }
        \\

        Batch size
        & 256
        & 512
        & 512
        & 512
        & 1024
        & 1024
        & 2048
        \\

        Training steps
        & 30K
        & 115K
        & 105K
        & 60K
        & 30K
        & 14K
        & 8K
        \\

        Warmup steps
        & --
        & 2000
        & 2000
        & 2000
        & 2000
        & 2000
        & 2000
        \\

        Loss weight ($\mathcal{L}_\mathrm{txt}$:$\mathcal{L}_\mathrm{gen}$)
        & $0 : 1$
        & $0.1 : 1$
        & $0.1 : 1$
        & $0.1 : 1$
        & $0.1 : 1$
        & $0.1 : 1$
        & $0.1 : 1$
        \\

        Sequence length
        & \multicolumn{7}{c}{8192}
        \\

        Generation resolution
        & $256^2$
        & $256^2 \rightarrow 512^2$
        & $512^2 \rightarrow 1024^2$
        & $512^2 \rightarrow 1024^2$
        & $1024^2 \rightarrow 2048^2$
        & $1024^2 \rightarrow 2048^2$
        & $1024^2 \rightarrow 2048^2$
        \\

        Understanding resolution
        & --
        & $512^2 \rightarrow 2048^2$
        & $512^2 \rightarrow 2048^2$
        & $512^2 \rightarrow 2048^2$
        & $512^2 \rightarrow 2048^2$
        & $512^2 \rightarrow 2048^2$
        & $512^2 \rightarrow 2048^2$
        \\



        \midrule

        \multicolumn{8}{l}{\textbf{Task sampling ratio}}
        \\

        Pure-text modeling
        & - & 0.10 & 0.10 & 0.10 & 0.10 & 0.05 & 0.05
        \\

        Image understanding
        & - & 0.30 & 0.30 & 0.30 & 0.30 & 0.15 & 0.20
        \\

        Image generation
        & 1.00 & 0.60 & 0.60 & 0.60 & 0.60 & 0.60 & 0.50
        \\

        Image editing
        & - & - & - & - & - & 0.20 & 0.25
        \\


        \bottomrule
    \end{tabular}%
    }

    \label{tab:hiuni_training_recipe}
\end{table*}

\begin{table}[t]
    \centering
    \small

    \caption{
        Configuration of the \emph{Pixel Diffusion Head} and \emph{HiGate}.
    }
    \label{tab:pixel_dit_higate_config}

    \resizebox{0.7\textwidth}{!}{%
    \begin{tabular}{l c}
        \toprule
        \textbf{Configuration} \\
        \midrule

        Number of Transformer blocks
        & $10$ \\

        Hidden dimension
        & $D=2560$ \\

        Attention
        & GQA with 32 query heads and 8 key-value heads \\

        Head dimension
        & $80$ \\

        Feed-forward network
        & SwiGLU, intermediate dimension $9728$ \\

        Flattened pixel dimension
        & $3\times32\times32=3072$ \\

        Pixel residual projection
        & $3072\rightarrow2560$ \\

        Semantic input
        & Diffusion-head hidden state $\mathbf{h}\in\mathbb{R}^{2560}$ \\

        Pixel input
        & Projected pixel residual $\mathbf{u}\in\mathbb{R}^{2560}$ \\

        Gate MLP
        & $2560\rightarrow2560\rightarrow1$ with SiLU activation \\







        Block-level AdaLN
        & Disabled \\

        \bottomrule
    \end{tabular}
    }
\end{table}

This section provides additional training details of \hiuni{}. Following the three-stage procedure described in Section~\ref{sec:experimental_setup}, training consists of generation warmup, joint pretraining, and supervised fine tuning. The complete training configuration is summarized in Table~\ref{tab:hiuni_training_recipe}.

\subsection{Training Stages}
\label{app:training_stages}

During generation warmup, we train the \emph{Pixel Diffusion Head} and \emph{HiGate} for 30K steps while keeping the pretrained visual encoder and backbone frozen. Only image generation data are used in this stage, with a fixed resolution of $256^2$. 

We then unfreeze the full model and perform joint pretraining for 310K steps on pure text modeling, image understanding, and image generation. The generation resolution is progressively increased from $256^2$ to $2048^2$, while the task sampling ratio remains $0.10{:}0.30{:}0.60$ throughout pretraining.

Finally, supervised fine tuning is performed for 22K steps. Image editing is introduced at this stage while the existing tasks are retained. The task mixture is adjusted between the two fine tuning phases as shown in Table~\ref{tab:hiuni_training_recipe}. Other phase specific settings, including batch size, learning rate schedule, and image resolution, are also provided in the table.

\subsection{Pixel Diffusion Head Configuration}
\label{app:pixel_dit_config}

The \emph{Pixel Diffusion Head} contains ten Transformer blocks with a hidden dimension of $2560$, matching the hidden dimension of the shared backbone. 
For each target visual token, the corresponding $32\times32$ RGB pixel patch is flattened and projected to the diffusion hidden dimension before being injected through \emph{HiGate}. The gate is predicted by a two layer MLP with SiLU activation. We do not apply an additional sigmoid or tanh activation to the gate output. 

\subsection{Optimization and Diffusion Settings}
\label{app:optimization_settings}

We use AdamW throughout training with $\beta_1=0.9$, $\beta_2=0.95$, and $\epsilon=10^{-8}$. For image editing, the reference and target images share the sequence budget with textual and special tokens, and their resolutions are adjusted jointly to keep the complete sequence within $8192$ tokens.
The relative loss weight between language modeling and diffusion is set to $0.1{:}1$, except for warmup.

For diffusion training, timesteps are sampled from a logit normal distribution with $\mu=-0.8$ and $\sigma=0.8$.
We use a resolution dependent noise scale according to the number of spatial tokens $N_g$.
Specifically, we sample $\mathbf{z}\sim\mathcal{N}(0,I)$ and define
\[
\boldsymbol{\epsilon}_{N_g}
=
\sigma_{N_g}\mathbf{z},
\qquad
\sigma_{N_g}
=
\sigma_0\sqrt{\frac{N_g}{N_0}},
\]
where $\sigma_0=1$, $N_0=64$, and $\sigma_{N_g}$ is clipped to $[1,8]$.
The noisy input and target velocity are
\[
\mathbf{x}_t
=
t\mathbf{x}_0+(1-t)\boldsymbol{\epsilon}_{N_g},
\qquad
\mathbf{v}
=
\mathbf{x}_0-\boldsymbol{\epsilon}_{N_g}.
\]
Since the diffusion head directly predicts the clean image $\hat{\mathbf{x}}_0$,
we convert it to the velocity prediction used by the training objective as
\[
\hat{\mathbf{v}}
=
\frac{\hat{\mathbf{x}}_0-\mathbf{x}_t}
{\max(1-t,0.05)}.
\]
The denominator is clipped only for numerical stability when $t$ approaches one and does not modify the forward interpolation or the target velocity.

The maximum multimodal sequence length is $8192$ tokens.
With an effective visual patch size of $32\times32$, an image of size $H\times W$ contributes approximately $(H/32)(W/32)$ visual tokens.




\subsection{Loss Aggregation}
\label{app:training_loss}

Language modeling and diffusion losses are normalized separately.
For pure text modeling and image understanding, cross entropy is averaged over all valid textual targets:
\[
\mathcal{L}_{\mathrm{txt}}
=
-\frac{1}{|\mathcal{T}|}
\sum_{(n,j)\in\mathcal{T}}
\log p_{\theta}
\left(
y_{n,j}
\mid
y_{n,<j},\mathbf{c}_n
\right),
\]
where $\mathcal{T}$ denotes all valid textual targets in the batch.

For image generation and editing, we first compute the mean squared velocity error independently for each image:
\[
\ell_{\mathrm{diff}}^{(n)}
=
\frac{1}{D_n}
\left\|
\hat{\mathbf{v}}_n-\mathbf{v}_n
\right\|_2^2,
\]
where $D_n$ denotes the number of elements in the velocity target, including all spatial positions and channels.
The diffusion objective is then averaged over all generation and editing examples:
\[
\mathcal{L}_{\mathrm{gen}}
=
\frac{1}{|\mathcal{I}_{\mathrm{gen}}|}
\sum_{n\in\mathcal{I}_{\mathrm{gen}}}
\ell_{\mathrm{diff}}^{(n)},
\]
where $\mathcal{I}_{\mathrm{gen}}$ contains both text to image generation and image editing examples.
This per image normalization prevents higher resolution samples from receiving larger effective weights, while the relative contribution of generation and editing follows the task sampling ratios in Table~\ref{tab:hiuni_training_recipe}.

\subsection{Training Dataset}
\label{app:data_composition}



Figure~\ref{fig:data_distribution} summarizes the composition of the training corpus used for image understanding, generation, and editing.
For image understanding, the data are mainly distributed across OCR and visual grounding, together with smaller portions of document, arts, sciences and agent-related data.
For image generation, the corpus is organized into three major groups, namely nature, people, and design.
For image editing, the data mainly cover subject-level, attribute-level, and structure-level edits.

The image understanding and generation corpora are constructed from proprietary datasets.
For image generation, most training data consist of real image-text pairs, with a small amount of synthesized images to control the style. For image understanding, we include training splits from some public datasets. Model-generated supervision is also used. Half of the pretraining data contain generated captions, mainly from Qwen series models, while $\sim97$\% of the SFT samples contain model-generated reasoning traces.
For image editing, the corpus contains about 4M examples collected from public datasets and prior work~\citep{ye2026imgedit,wei2025omniedit,bai2024humanedit,
chen2026scaleedit,chen2026vins,wu2025omnigen2,hu2024vivid}, together with additional private editing pairs produced by our internal data-generation pipeline.
For ImgEdit, only the official training split is included in the training corpus, while evaluation is conducted exclusively on its official test split.

\section{Evaluation Details}
\label{app:benchmark_inference}

\subsection{Image Understanding}
\label{app:understanding_inference}

We evaluate \hiuni{} on the image understanding benchmarks reported in Table~\ref{tab:understanding_main}. 
We follow the official benchmark splits and scoring protocols, with benchmark specific answer extraction when required.
For the standard image understanding benchmarks, we use $\texttt{min\_pixels}=262144$, $\texttt{max\_pixels}=4194304$, and $\texttt{max\_model\_len}=16384$. 
For evaluations with reasoning mode, the model first generates its reasoning process and then produces the final answer. 
Only the final answer is passed to the benchmark scorer. 
If a valid answer is not completed in the first generation, decoding is continued from the existing context without restarting the example.

We follow the official benchmark splits and scoring protocols for all reported evaluations.

\paragraph{MathVista.}
We evaluate on the official MathVista-Mini split and follow its standard evaluation protocol and scorer. 
The reported score is averaged over two complete evaluations.
\paragraph{MMBench.}
We evaluate MMBench-DEV-EN using the official circular evaluation protocol and report the corresponding overall score.
\paragraph{RealWorldQA.}
We evaluate the official RealWorldQA set and report the EM Overall score.
\paragraph{MMStar.}
We evaluate the official MMStar set and report the Overall score across all categories.
\paragraph{HallusionBench.}
We follow the official evaluation protocol and report its Overall score.
\paragraph{AI2D.}
We evaluate on AI2D-Test and use the official EM protocol for multiple choice prediction.
\paragraph{OCRBench.}
We evaluate OCRBench v1 and report the official Final Score as a percentage.
\paragraph{OmniDocBench.}
We evaluate the English split of OmniDocBench v1.0 and report the official OverallEdit score, where a lower value indicates better performance.

\subsection{Image Generation}
\label{app:generation_inference}

We use the original benchmark prompts without prompt rewriting or post processing.
We use the official evaluators. 
The inference settings are listed below.

\paragraph{Qwen-Image-Bench.}
We use a CFG scale of 6, a shift of 5, and 100 sampling steps, and generate one image for each prompt. 

\paragraph{OneIG.}
We use a CFG scale of 6, a shift of 4, and 25 sampling steps, and generate four images for each prompt. 

\paragraph{LongText.}
We use a CFG scale of 8, a shift of 8, and 100 sampling steps, and generate four images for each prompt.

\begin{table*}[t]
    \centering
    \caption{
        Detailed results on OneIG.
    }
    \label{tab:oneig_results}

    \scriptsize
    \resizebox{\textwidth}{!}{%
    \begin{tabular}{
        l
        c |
        *{6}{c} | *{6}{c}
    }
        \toprule

        \multicolumn{1}{l}{\multirow{3}{*}{\textbf{Models}}}
        & \multicolumn{1}{c|}{\multirow{3}{*}{\textbf{Size}}}
        & \multicolumn{6}{c|}{\textbf{OneIG-ZH}}
        & \multicolumn{6}{c}{\textbf{OneIG-EN}}
        \\

        \cmidrule(lr){3-8}
        \cmidrule(lr){9-14}

        &
        & Alignment
        & Text
        & Reasoning
        & Style
        & Diversity
        & Overall
        & Alignment
        & Text
        & Reasoning
        & Style
        & Diversity
        & Overall
        \\

        \midrule

        \rowcolor{gray!15}
        \multicolumn{14}{c}{
            \textit{Generation-only Models}
        }
        \\

        GLM-Image$^\star$
        & 9B + 7B
        & 0.738 & 0.976 & 0.284 & 0.335 & 0.221 & 0.511
        & 0.805 & 0.969 & 0.298 & 0.353 & 0.213 & 0.528
        \\

        Qwen-Image$^\star$
        & 20B
        & 0.825 & 0.963 & 0.267 & 0.405 & 0.279 & 0.548
        & 0.882 & 0.891 & 0.306 & 0.418 & 0.197 & 0.539
        \\

        
        ERNIE-Image$^\star$
        & 8B
        & 0.842 & 0.898 & 0.266 & 0.421 & 0.177 & 0.521
        & 0.891 & 0.967 & 0.295 & 0.447 & 0.169 & 0.554
        \\
        
        \midrule

        \rowcolor{gray!15}
        \multicolumn{14}{c}{
            \textit{Unified Multimodal Models}
        }
        \\

        Bagel
        & 7B + 7B
        & 0.672 & 0.365 & 0.186 & 0.357
        & \textbf{0.268} & 0.370
        & 0.769 & 0.244 & 0.173 & 0.367
        & \textbf{0.251} & 0.361
        \\

        InternVL-U$^\star$
        & 2B + 1.7B
        & 0.750 & 0.900 & 0.230 & 0.370
        & \underline{0.260} & 0.500
        & 0.820 & 0.740 & 0.270 & 0.400
        & \underline{0.250} & 0.500
        \\

        SenseNova-U1$^\star$
        & 8B + 8B
        & \textbf{0.826} & \textbf{0.977} & \textbf{0.303}
        & 0.392 & 0.176 & \textbf{0.535}
        & \textbf{0.882} & \textbf{0.969} & \textbf{0.330}
        & 0.396 & 0.166 & \underline{0.549}
        \\

        UniSpace$^\star$
        & 8B + 8B
        & 0.807 & 0.881 & \underline{0.276} & \textbf{0.455}
        & 0.244 & \underline{0.533}
        & \underline{0.860} & 0.937 & \underline{0.311}
        & \textbf{0.467} & 0.233 & \textbf{0.561}
        \\

        \midrule

        \rowcolor{blue!8}
        \textbf{\hiuni}
        & 4B + 0.9B
        & 0.795 & \underline{0.953} & 0.259 & \underline{0.426}
        & 0.185 & 0.524
        & 0.851 & \underline{0.941} & 0.275 & \underline{0.443}
        & 0.179 & 0.538
        \\

        \bottomrule
    \end{tabular}%
    }
\end{table*}

\subsection{Image Editing}
\label{app:editing_inference}

We evaluate \hiuni{} on ImgEdit and GEdit using the same image editing inference pipeline. 
For both benchmarks, we generate one image for each instruction using $50$ sampling steps with the Euler scheduler, a shift of $3.0$, and a CFG scale of $4.0$. 
No additional post processing is applied.

\paragraph{ImgEdit.}
We follow the official evaluation protocol and prompts, using GPT-5.2 as the judge model.
Each edited image is scored on a $0$ to $5$ scale, and the Overall score is averaged across edit types.

\paragraph{GEdit.}
We follow the VIEScore evaluation protocol and use GPT-5.2 as the judge model. 
Each result is evaluated on semantic consistency (G\_SC) and perceptual quality (G\_PQ), both on a $0$ to $10$ scale, with their geometric mean reported as G\_O. 
Scores are first averaged within each edit type and then across all edit types.

\begin{table*}[t]
\centering
\caption{
Detailed results on GEdit.}
\label{tab:gedit_detail}
\begin{subtable}[t]{\textwidth}
\centering
\label{tab:gedit_detail_cn}
\scriptsize
\begin{adjustbox}{max width=\textwidth}
\begin{tabular}{@{}l c | w{c}{4.1em} *{10}{w{c}{3.05em}} w{c}{3.2em}@{}}
        \toprule
        \multicolumn{1}{l}{\multirow{3}{*}{\textbf{Models}}}
        & \multicolumn{1}{c|}{\multirow{3}{*}{\textbf{Size}}}
        & \multicolumn{11}{c}{\textbf{Chinese prompts}} & \\
        \cmidrule(lr){3-14}
        &
        & background & color & material & motion & human & style & add & remove & replace & text & tone & \textbf{Avg.} \\
        \midrule

\rowcolor{gray!10}
\multicolumn{14}{c}{\textit{Editing-only Models}}
\\
Qwen-Image-2.1
& 7B
& 6.83 & 7.26 & 6.75 & 6.73 & 6.04 & 6.88 & 7.51 & 4.29 & 6.87 & 7.66 & 7.22
& 6.73
\\

LongCat-Image-Edit
& 6B
& 7.25 & 7.41 & 6.97 & 7.20 & 6.39 & 6.80 & 7.53 & 7.51 & 7.34 & 7.32 & 6.92
& 7.15
\\

Qwen-Image-Edit-2511
& 20B
& 7.01 & 7.59 & 6.79 & 7.23 & 6.66 & 6.85 & 7.54 & 7.48 & 7.28 & 7.65 & 7.00
& 7.19
\\

\midrule

\rowcolor{gray!10}
\multicolumn{14}{c}{\textit{Unified Multimodal Models}}
\\
InternVL-U
& 2B + 1.7B
& 6.55 & 5.66 & 5.32 & 5.92 & 5.42 & 6.15 & 6.12 & 5.31 & 6.25 & 4.55 & 5.87
& 5.74
\\

Bagel
& 7B + 7B
& 6.26 & 6.59 & 5.97 & 6.26 & 5.44 & 5.83 & 7.05 & 6.99 & 6.83 & 6.30 & 5.91
& 6.31
\\

SenseNova-U1
& 8B + 8B
& \textbf{7.01} & \textbf{7.28} & \textbf{6.95} & \textbf{7.23} & \textbf{6.33} & \textbf{6.81} & \textbf{7.39} & \textbf{7.17} & \underline{7.01} & \textbf{7.53} & \textbf{6.89}
& \textbf{7.05}
\\

UniSpace
& 8B + 8B
& 6.62 & 7.00 & \underline{6.40} & \underline{6.84} & \underline{6.05} & \underline{6.73} & 5.77 & \underline{7.15} & \textbf{7.03} & 6.26 & \underline{6.44}
& 6.57
\\

\midrule

\rowcolor{blue!8}
\textbf{\hiuni}
& 4B + 0.9B
& \underline{6.65} & \underline{7.03} & \underline{6.40} & 6.80 & 6.01 & 6.42 & \underline{7.08} & 6.92 & 6.92 & \underline{7.28} & 6.18
& \multicolumn{1}{w{c}{3.2em}}{\underline{6.70}}
\\
        \bottomrule
\end{tabular}
\end{adjustbox}
\end{subtable}


\begin{subtable}[t]{\textwidth}
\centering
\label{tab:gedit_detail_en}
\scriptsize
\begin{adjustbox}{max width=\textwidth}
\begin{tabular}{@{}l c | w{c}{4.1em} *{10}{w{c}{3.05em}} w{c}{3.2em}@{}}
        \toprule
        \multicolumn{1}{l}{\multirow{3}{*}{\textbf{Models}}}
        & \multicolumn{1}{c|}{\multirow{3}{*}{\textbf{Size}}}
        & \multicolumn{11}{c}{\textbf{English prompts}} & \\
        \cmidrule(lr){3-14}
        &
        & background & color & material & motion & human & style & add & remove & replace & text & tone & \textbf{Avg.} \\
        \midrule

\rowcolor{gray!10}
\multicolumn{14}{c}{\textit{Editing-only Models}}
\\
Qwen-Image-2.1
& 7B
& 7.08 & 7.40 & 6.82 & 7.00 & 6.25 & 6.57 & 7.43 & 5.31 & 7.25 & 7.82 & 7.24
& 6.92
\\

LongCat-Image-Edit
& 6B
& 7.20 & 7.43 & 6.72 & 7.20 & 6.28 & 6.86 & 7.44 & 7.34 & 7.20 & 7.37 & 7.02
& 7.10
\\

Qwen-Image-Edit-2511
& 20B
& 7.12 & 7.53 & 6.85 & 7.42 & 6.53 & 6.93 & 7.47 & 7.34 & 7.19 & 7.73 & 6.80
& 7.17
\\

\midrule

\rowcolor{gray!10}
\multicolumn{14}{c}{\textit{Unified Multimodal Models}}
\\
InternVL-U
& 2B + 1.7B
& 6.29 & 5.86 & 5.44 & 6.61 & 5.21 & 6.35 & 6.32 & 5.03 & 6.18 & 4.65 & 5.25
& 5.74
\\

Bagel
& 7B + 7B
& 6.28 & 6.49 & 5.63 & 6.51 & 5.32 & 5.89 & 7.03 & 6.67 & 6.87 & 6.48 & 5.77
& 6.27
\\

SenseNova-U1
& 8B + 8B
& \textbf{7.00} & \underline{6.94} & \underline{6.59} & \textbf{7.05} & 6.05 & \textbf{6.67} & \textbf{7.28} & \textbf{7.25} & \underline{7.04} & \textbf{7.55} & \textbf{6.86}
& \textbf{6.94}
\\

UniSpace
& 8B + 8B
& 6.63 & 6.78 & \textbf{6.63} & \textbf{7.05} & \textbf{6.17} & 6.57 & 6.61 & 6.88 & \textbf{7.16} & 6.30 & \underline{6.70}
& 6.68
\\

\midrule

\rowcolor{blue!8}
\textbf{\hiuni}
& 4B + 0.9B
& \underline{6.80} & \textbf{7.14} & 6.43 & \underline{6.98} & \underline{6.10} & \underline{6.58} & \underline{7.16} & \underline{7.02} & 6.98 & \underline{7.21} & 6.57
& \multicolumn{1}{w{c}{3.2em}}{\underline{6.81}}
\\
        \bottomrule
\end{tabular}
\end{adjustbox}
\end{subtable}
\end{table*}

\begin{figure}[t]
    \centering
    \includegraphics[width=0.75\linewidth]{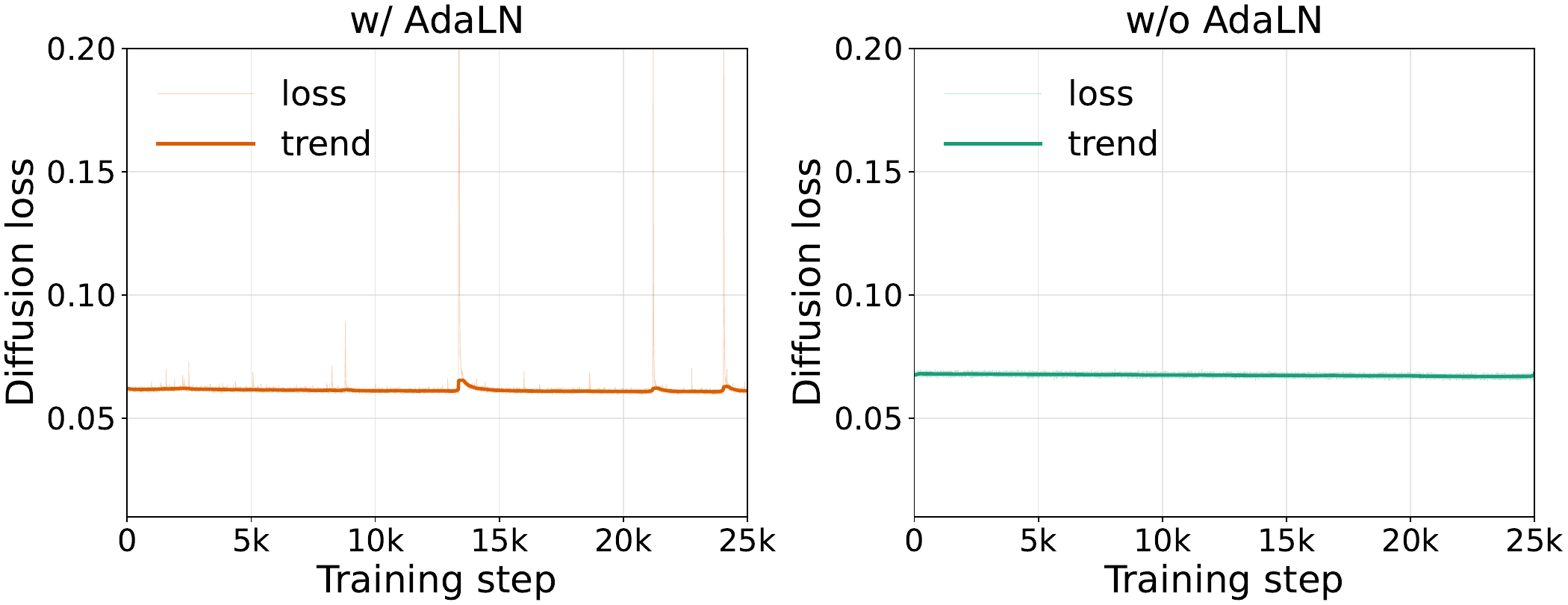}
    \caption{
    Diffusion training trajectories  from development runs with and without AdaLN in the Pixel Diffusion Head. 
    The run without AdaLN (right) shows fewer loss spikes.
    The two curves are collected at different development stages and therefore provide only a qualitative comparison rather than a controlled isolation of the effect of AdaLN.
    }
    \label{fig:adaln_ablation}
\end{figure}

\section{Optimization Stability of the Pixel Diffusion Head}
\label{app:adaln_ablation}

During early development, we observed that the \emph{Pixel Diffusion Head} with AdaLN frequently produced abrupt loss spikes when trained directly in pixel space.
We therefore simplified the diffusion blocks by removing repeated AdaLN modulation while retaining timestep conditioning through the shared backbone and the final prediction layer.
Figure~\ref{fig:adaln_ablation} shows representative training trajectories.
These two runs use different resolution schedules and come from different development stages.
The curve without AdaLN exhibits a noticeably smoother optimization trajectory with fewer large fluctuations, consistent with the training behavior observed in our subsequent experiments.



\section{Extending to Chain-of-Thought Image Generation}
\label{app:cot_generation}

\paragraph{Formulation}
As an additional experiment, we extend \hiuni{} to chain-of-thought (CoT) based image generation, where the model first produces an intermediate textual reasoning sequence before generating the target image. 
Given a text prompt $\mathbf{p}$, the model autoregressively generates a textual reasoning sequence $\mathbf{r}$, which is then included in the conditioning context as $\mathbf{C}_{\mathrm{cot}}=(\mathbf{p},\mathbf{r})$.
Image generation follows the same formulation as in Section~\ref{sec:training},
$
\hat{\mathbf{x}}_0 = f_{\theta}\!\left(
\mathbf{x}_t,t,\mathbf{C}_{\mathrm{cot}}
\right).
$
The reasoning sequence is optimized with the language modeling objective $\mathcal{L}_{\mathrm{txt}}$, while the target image is optimized with the same diffusion objective $\mathcal{L}_{\mathrm{gen}}$ used for standard image generation.

\paragraph{Training}
We obtain the CoT model by continuing training from a \hiuni{} checkpoint after Phase III of joint pretraining, where the training data are derived from UniReason~\citep{wang2026unireason}.
The model is further trained for 10K steps with a global batch size of 512.
Within the image generation data, standard text-to-image examples and CoT generation examples are sampled with a ratio of $1{:}1$, while image understanding and pure-text modeling data are retained during training.
No additional backbone or generation module is introduced.


\paragraph{Results}
We evaluate the extension on the WISE benchmark~\citep{niu2025wise}, with results reported in Table~\ref{tab:wise}.
\hiuni$_\mathrm{w/ cot}$ achieves an Overall score of $0.75$, exceeding DeepGen 1.0 and Bagel with CoT and remaining competitive with stronger recent image generation models.
It performs particularly well on Cultural, Time, and Chemistry, with scores of $0.81$, $0.68$, and $0.78$, respectively.
These results are obtained through a short fine-tuning stage without modifying the shared backbone or generation pathway, indicating that the \hiuni{} architecture can be readily extended to reasoning based image generation.
In future work, we plan to integrate this capability into the main training pipeline and further study its interaction with standard image generation.

\begin{table}[t]
\centering


\caption{Results on the WISE benchmark for chain of thought image generation.}
\label{tab:wise}

\centering
\scriptsize
\resizebox{0.85\textwidth}{!}{%

\resizebox{\linewidth}{!}{%
\begin{tabular}{@{}lccccccc@{}}
\toprule
\textbf{Model}
& \textbf{Overall}
& \makecell{\textbf{Cultural}}
& \makecell{\textbf{Time}}
& \makecell{\textbf{Space}}
& \makecell{\textbf{Biology}}
& \makecell{\textbf{Physics}}
& \makecell{\textbf{Chemistry}} \\
\midrule

DeepGen 1.0$^\star$
& 0.57 & 0.65 & 0.41 & 0.72 & 0.39 & 0.59 & 0.45 \\

Bagel$_\mathrm{w/ cot}$$^\star$
& 0.63 & 0.78 & 0.63 & 0.57 & 0.38 & 0.55 & 0.51 \\

SenseNova-U1-8B$^\star$
& 0.68 & 0.69 & 0.60 & 0.74 & 0.63 & 0.72 & 0.69 \\

SenseNova-U1-8B$_\mathrm{w/ cot}$$^\star$
& 0.79 & 0.80 & 0.74 & 0.86 & 0.76 & 0.78 & 0.80 \\

Qwen-Image-2.0$^\star$
& 0.80 & 0.82 & 0.65 & 0.90 & 0.79 & 0.80 & 0.75 \\

GPT-Image-1.5$^\star$
& 0.83 & 0.89 & 0.69 & 0.88 & 0.80 & 0.76 & 0.78 \\


\rowcolor{blue!8}
\hiuni$_\mathrm{w/ cot}$
& 0.75 & 0.81 & 0.68 & 0.78 & 0.63 & 0.74 & 0.78 \\

\bottomrule
\end{tabular}%
}
}
\end{table}

\begin{table}[t]
    \centering
    \caption{
        Effect of image-editing supervision on image understanding and generation.
        All other settings are kept fixed, with image-editing data as the only
        changed factor.
    }
    \label{tab:editing_to_understanding}

    \scriptsize
    \resizebox{\textwidth}{!}{%
    \begin{tabular}{l *{7}{c} c}
        \toprule

        \multicolumn{1}{l}{\multirow{3}{*}{\textbf{Configuration}}}

        & \multicolumn{7}{c}{\textbf{Understanding}}
        & \multicolumn{1}{c}{\textbf{Generation}}
        \\

        \cmidrule(lr){2-8}
        \cmidrule(lr){9-9}

        & \textbf{MMBench$_{\mathrm{EN}}$}
        & \textbf{RealWorldQA}
        & \textbf{MMStar}
        & \textbf{HallusionBench}
        & \textbf{AI2D}
        & \textbf{OCRBench}
        & \textbf{Avg.}
        & \textbf{OneIG Avg.}
        \\

        \midrule

        w/o Editing Supervision
        & 69.58
        & 64.58
        & 50.20
        & 48.43
        & 72.54
        & 70.50
        & 62.64
        & 0.538
        \\

        \rowcolor{blue!8}
        w/ Editing Supervision
        & 72.29
        & 66.80
        & 55.00
        & 43.83
        & 79.24
        & 69.80
        & 64.49
        & 0.528
        \\

        \bottomrule
    \end{tabular}%
    }
\end{table}

\section{Effect of Editing Supervision}
\label{app:editing_supervision}

We further examine cross-task interaction by comparing two training settings that differ only in the inclusion of image-editing supervision.
We keep the total training token budget, model architecture, and optimization settings fixed, while adjusting the task sampling ratios to include image editing. The underlying understanding and generation data sources remain unchanged.

As shown in Table~\ref{tab:editing_to_understanding}, editing supervision improves the average understanding score from $62.64$ to $64.49$, with gains on four of the six evaluated benchmarks.
The improvements are particularly pronounced on AI2D-Test and MMStar, which increase by $6.70$ and $4.80$ points, respectively, indicating that the transferred supervision may be particularly useful for structured visual reasoning.
In contrast, HallusionBench decreases by $4.60$ points, while OCRBench changes only slightly.
This non-uniform pattern suggests that the benefit of editing supervision is task dependent rather than a general improvement across all understanding capabilities.
At the same time, the average OneIG score slightly decreases from $0.538$ to $0.528$.
Together with the understanding results, this suggests that editing supervision introduces both positive transfer and mild interference under shared training.
It can improve several understanding capabilities, while also changing the optimization balance among tasks rather than uniformly benefiting all of them.


\section{Qualitative Results}
\label{app:qualitative_results}

\subsection{Image Generation}
\begin{figure}
    \centering
    \includegraphics[width=1.0\linewidth]{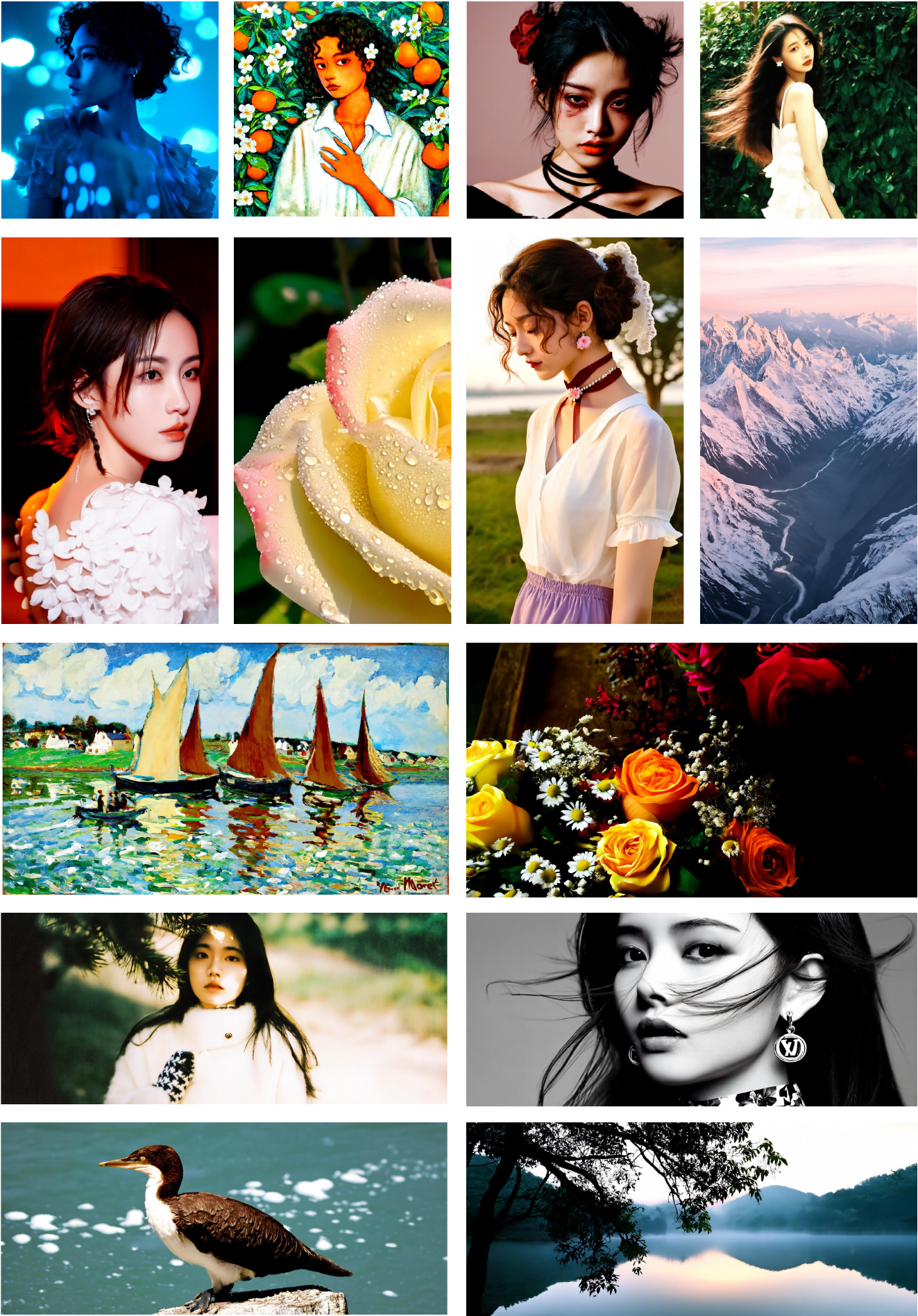}
    \caption{Showcase of \hiuni{} in image generation.}
    \label{fig:t2i_showcase}
\end{figure}

\begin{figure}
    \centering
    \includegraphics[width=1.0\linewidth]{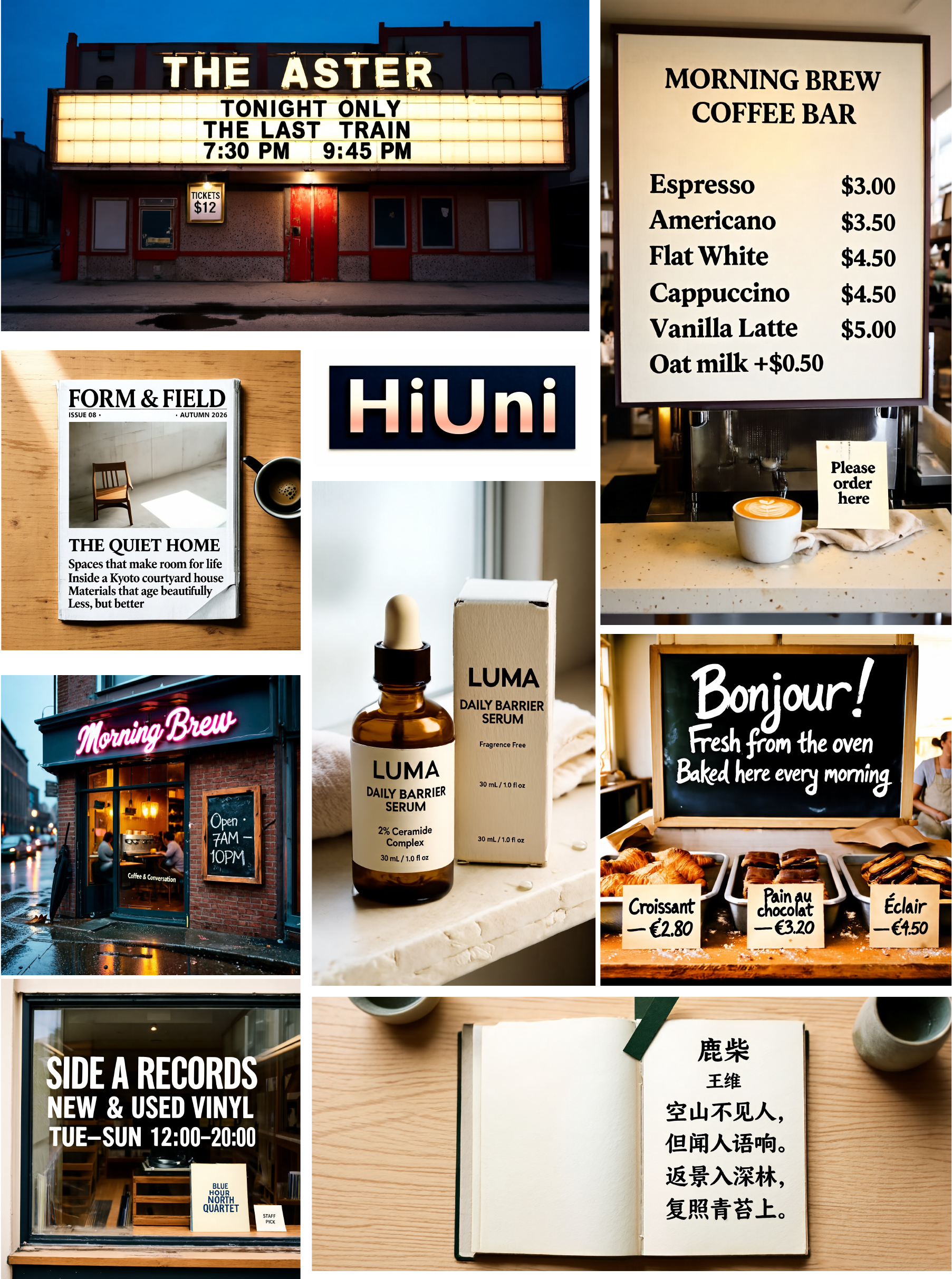}
    \caption{Showcase of \hiuni{} in text rendering.}
    \label{fig:textrender}
\end{figure}

Representative image generation results of \hiuni{} are shown in Figure~\ref{fig:t2i_showcase}, and text rendering results of \hiuni{} are shown in Figure~\ref{fig:textrender}.

\subsection{Image Editing}
\begin{figure}
    \centering
    \includegraphics[width=1.0\linewidth]{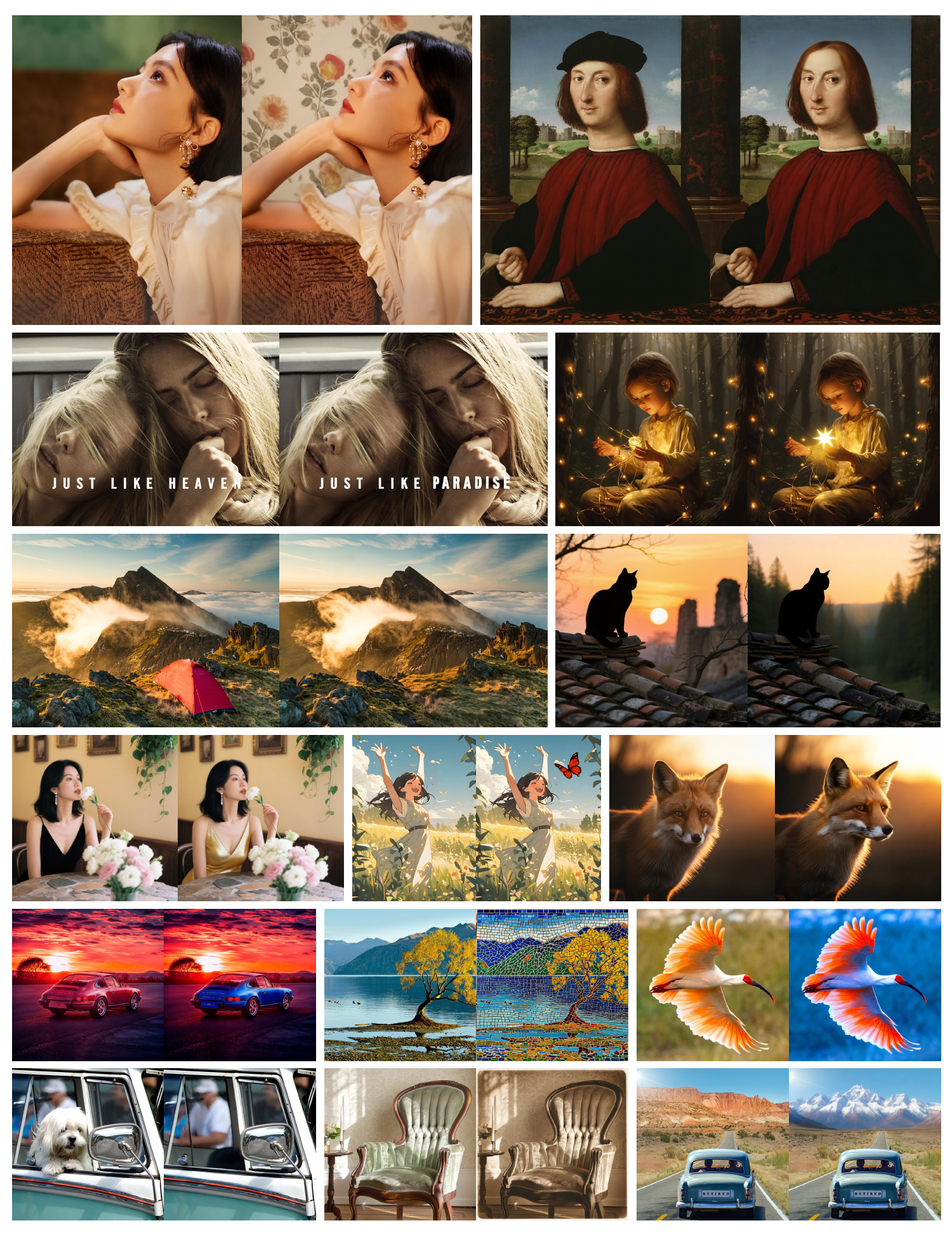}
    \caption{Showcase of \hiuni{} in image editing.}
    \label{fig:edit_showcase}
\end{figure}

Representative image editing results of \hiuni{} are shown in Figure~\ref{fig:edit_showcase}, covering several common editing types.

\end{document}